%% file: main.tex
\documentclass[10pt,twocolumn,letterpaper]{article}

\usepackage{iccv}
\usepackage{times}
\usepackage{epsfig}
\usepackage{graphicx}
\usepackage{amsmath}
\usepackage{amssymb}
\usepackage{tikz}
\usepackage{wrapfig}
\usepackage{subcaption}
\usepackage{booktabs}
\usepackage[accsupp]{axessibility}

\newtheorem{theorem}{Theorem}[section]

\usepackage[pagebackref=true,breaklinks=true,letterpaper=true,colorlinks,bookmarks=false]{hyperref}

\iccvfinalcopy % *** Uncomment this line for the final submission

\def\iccvPaperID{11301} % *** Enter the ICCV Paper ID here

\ificcvfinal\fi

\begin{document}

%%%%%%%%% TITLE
\title{Stable and Causal Inference for Discriminative Self-supervised Deep Visual Representations}

\author{Yuewei Yang, Hai Li, Yiran Chen\\
Duke University\\
Durham, USA\\
{\tt\small yuewei.yang@duke.edu, hai.li@duke.edu, yiran.chen@duke.edu}
% For a paper whose authors are all at the same institution,
% omit the following lines up until the closing ``}''.
% Additional authors and addresses can be added with ``\and'',
% just like the second author.
% To save space, use either the email address or home page, not both
% \and
% Second Author\\
% Institution2\\
% First line of institution2 address\\
% {\tt\small secondauthor@i2.org}
}

\maketitle
% Remove page # from the first page of camera-ready.
\ificcvfinal\thispagestyle{empty}\fi

%%%%%%%%% ABSTRACT
\begin{abstract}
   In recent years, discriminative self-supervised methods have made significant strides in advancing various visual tasks. The central idea of learning a data encoder that is robust to data distortions/augmentations is straightforward yet highly effective. Although many studies have demonstrated the empirical success of various learning methods, the resulting learned representations can exhibit instability and hinder downstream performance. In this study, we analyze discriminative self-supervised methods from a causal perspective to explain these unstable behaviors and propose solutions to overcome them. Our approach draws inspiration from prior works that empirically demonstrate the ability of discriminative self-supervised methods to demix ground truth causal sources to some extent. Unlike previous work on causality-empowered representation learning, we do not apply our solutions during the training process but rather during the inference process to improve time efficiency. Through experiments on both controlled image datasets and realistic image datasets, we show that our proposed solutions, which involve tempering a linear transformation with controlled synthetic data, are effective in addressing these issues.
\end{abstract}

%%%%%%%%% BODY TEXT
\section{Introduction}

\begin{figure}
    \centering
    \resizebox{0.35\textwidth}{!}{
    \input{images/problem.tikz}}
    \caption{During the training of discriminative SSL, aligning positive representations will be robust to the changes applied as augmentations (\textcolor{red}{red arrows}). However, during the inference stage, one small change in the data variable (such as \textit{view angle}) will result in an unexpected degradation on downstream performance.}
    \label{fig:problem}
\end{figure}
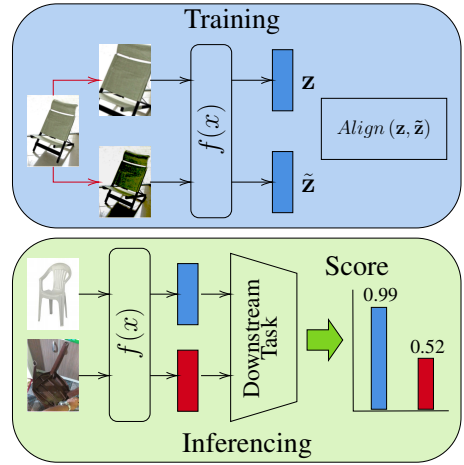

Learning generalized representation with unlabeled data is a challenging task in various fields, but Self-Supervised Learning (\textbf{SSL}) has recently demonstrated remarkable success in learning semantic invariant representations without labels \cite{jaiswal2020survey,jing2020self,liu2021self}. There are two main types of self-supervised learning (SSL) based on the pretext task used: generative and discriminative SSL, with generative SSL reconstructing altered or distorted data to its original input \cite{chen2020generative,he2022masked,hinton1993autoencoders,pathak2016context,vincent2010stacked,wei2022masked}and early discriminative SSL predicting easily designed labels and task-specific representations that are not very generalizable \cite{gidaris2018unsupervised,noroozi2016unsupervised,zhang2016colorful}. More recent discriminative SSL trains the model to identify similarities and differences between pairs of augmented examples \cite{caron2020unsupervised,chen2020simple,chen2021exploring,grill2020bootstrap,he2020momentum,zbontar2021barlow}. The success of SSL in deep image models has resulted in progress in other data modalities \cite{liu2021self,liu2022audio,liu2022graph,schiappa2022self,shurrab2022self} and attention-based models like transformers \cite{chen2021empirical,caron2021emerging,Li2021,Xie2021}. Recent discriminative SSL aims to learn content and semantic invariant representations that are robust to data augmentations, but the learned representations can be \textit{unstable} when one subtle factor of the data is changed to a value that is not accessible through all augmentations. To avoid the high cost of incorporating all possible subtle changes during training, insights are needed to uncover the root cause of instability and find a solution to prevent performance deterioration during inference. Figure \ref{fig:problem} summarizes this deterioration effect.

Causality \cite{Pearl2011} is a vital tool to investigate the causal relationships between variables in observational data, and can uncover the underlying causal factors that explain unexpected model behavior due to changes in the environment. Independent Component Analysis (\textbf{ICA}) is often used to disentangle sources in unsupervised training \cite{Hyvarinen2016,Hyvarinen2017,Hyvarinen2020,Khemakhem2019,Klindt2020}, and causal analysis has been applied in follow-up works \cite{Zimmermann2021,Von2021} to examine the empirical success of SSL under an ICA framework. However, while these works identify factors that contribute to SSL's success, they do not address the problem's unstable mode, which can cause a severe performance drop when underlying factors shift slightly to an unseen environment. Some works \cite{Mitrovic2020,HaoChen2021,Lee2021} attempt to incorporate causality during the training process to identify and alleviate the impact of such shifts, but this approach is time-costly and only marginally improves performance compared to non-causal SSL methods. A more time-efficient and accessible solution would be to simply reverse the unstable shift during inference.

We aim to address the issue of unstable behavior during the \textbf{inference stage} by building upon previous theories of successful InfoNCE-facilitated contrastive SSL and extending it to \textbf{all recent SSL methods} with additional assumptions and constraints. Drawing inspiration from the relationship between the ground truth positive pairs distribution and learned positive pairs distribution, we demonstrate that the approximated transformation between the ground truth representations and learned representations is orthogonal to the augmentations applied during training. However, a change in the data factor/variable that violates the conditions for successful SSL can cause a corresponding shift in the inferred representation, resulting in a decline in downstream performance. This change of data factor/variable can be a change in the background, texture, or view angles etc. To overcome this issue, we propose learning targeted transformations that regularize the violating shift and restore performance on the unseen data shift. This approach effectively avoids the undesirable behavior and improves performance on previously unseen data shifts.

To summarize, our contributions are following:
\begin{itemize}
    \item Through the use of a comparable data generation process in prior research, we show that \textbf{All} current SSL techniques benefit from the alignment of positive pairs.
    \item Through our alternative derivation of the transformation matrix between the ground truth representation and the learned representation, we have shown that the augmentations applied during training are orthogonal to the resulting matrix.
    \item By interpreting a change in the data variable causally, we propose two solutions \textbf{focusing on inference} to modify the negative shift in representation space caused by such a change.
    \item We validate the proposed solutions by conducting experiments on both controlled and realistic datasets, providing evidence for their efficacy during the inference stage without retraining the pretrained models.
\end{itemize}

\section{Related Work}
\textbf{Discriminative SSL} learns invariant representations from positive pairs of unlabeled data samples, but previous attempts to use trivial labels like colors\cite{zhang2016colorful}, rotations\cite{gidaris2018unsupervised}, and patch positions\cite{noroozi2016unsupervised} offer minimal benefits for complex downstream tasks due to their easy augmentations. Recent discriminative SSL apply random augmentations to generate two views of an image sample and train an encoder to extract representations for maximizing similarity between the paired augmentations. SimCLR and MoCo\cite{chen2020simple,he2020momentum} are pioneer SSL works that maximize the cosine similariy between positive pairs and minimize the cosine similarity between negative pairs via optimizing the InfoNCE loss\cite{Oord}. Immense resources are used to enforce a large number of negative samples since a larger number can tighten the upper bound of the mutual information between positive pairs\cite{Oord}. Later advancement of discriminative SSL excludes the notion of negative pairs by only aligning positive pairs and preventing \textit{representation collapse} through various regularizations. BYOL\cite{grill2020bootstrap} predicts an Exponential-Moving-Averaged (EMA) representation of one view with a projected representation of another view. SimSiam\cite{chen2021exploring} maximizes similarity between a projected representation and a detached representation of two positive samples. Unlike previous work focusing optimization on an instance level, Barlow Twins\cite{zbontar2021barlow} encourages high similarity in corresponding feature dimensions and discourages redundancy across different feature dimensions between two views of a data sample. Detailed formulations are exhibited in \ref{appendix:ssl formulations}.

Mutual Information is a different perspective on the behaviour of discriminative SSL. Referred to InfoMax principle\cite{linsker1988self}, the MI between different transformations of a data sample is maximized via optimizing the InfoNCE loss\cite{bachman2019learning,hjelm2018learning,khosla2020supervised}. Though showing theoretical relation between optimizing InfoNCE and maximizing MI between positive pairs, the underlying factors instructing the behaviour of different SSL methods are not explored. Nonlinear ICA\cite{hyvarinen1999nonlinear}, on the other hand, captures complex data structures of SSL methods by disentangling underlying factors via minimizing the mutual information between learned representations and the original data input\cite{dinh2014nice,xin2021disentangled,erdogmus2004minimax}. Other works associate the nonlinear ICA objectives with the contrastive SSL so the MI between positive pairs are maximized and negative pairs are minimized.

Other researchers have explored \textbf{causality and causal inference} as a means of understanding the success of discriminative SSL. Prior work has focused on partitioning the InfoNCE loss to \textit{alignment} between the positive pairs and \textit{uniformity} between aggregations of all positive clusters\cite{Wang2020}. By formulating a data generation process, \cite{Zimmermann2021} empirically explains that networks optimized via InfoNCE infer an orthogonal transformation of the ground truth latent representations. Furthermore \cite{Von2021} validates that that augmentations used in both generative and discriminative SSL isolate the \textit{content} factor from the \textit{style} factor.  Our theory and work draw great inspirations from these two work. However, instead of solely focusing on InfoNCE-driven SSL and two factors, this work extends the framework to all recent discriminative SSL methods and identifies reasons for unstable circumstances analytically. We also propose methods to nullify the negative effects of unstable representations during inference.

\textbf{Domain Adaptation} is a strategy to bridge the gap between the model performance on a source domain and that on a target domain \cite{ben2006analysis,wang2018deep}. Feature adaptation methods try to learn a new feature representation that is more invariant to the domain shift\cite{volpi2018adversarial,duan2012learning,daume2009frustratingly,li2013learning}, while instance adaptation methods try to reweight the importance of the labeled source examples to better align with the target domain\cite{long2014transfer,cao2018unsupervised,hsu2015unsupervised}. Recent studies also implement a contrastive framework to learn a shared latent space between the source and target domains by maximizing the agreement between representations of corresponding samples while minimizing the agreement between representations of non-corresponding samples \cite{wang2022cross,thota2021contrastive,kang2019contrastive}. Unlike these works focusing on adapting to a target domain for better performance, we investigate the underlying causal factors for the performance gap and based on the findings we propose easy solutions to connect the gap.

\section{Theory}
In this section, we build on previous data generation process \cite{Zimmermann2021,Von2021} (\ref{formulation}) and show that current SSL techniques benefit from the alignment of positive pairs (\ref{alignment and regularization}). During a deeper dive into the generation process, we show the relation between the ground truth representations and the inferred representations, and this relation is an linear transformation matrix that is orthogonal to the augmentations applied during training (\ref{augmentation orthogonal}). Finally, we disclose the causal reason for the unstable behaviour caused by a change in the data variable during the inference stage and propose analytical solutions to address this unstable issue (\ref{solution}).

\subsection{Problem formulation}\label{formulation}
% \setlength{\intextsep}{0pt}%
% % \setlength{\columnsep}{5pt}%
% \begin{wrapfigure}{r}{0.23\textwidth} 
% \centering
% \resizebox{!}{0.15\textwidth}{
%     \input{images/generation_causal.tikz}}
%     \caption{A causal graph for data generation process.} \label{fig:generation_causal}
% \end{wrapfigure} 
\begin{figure}[htbp]
    \centering
    \resizebox{!}{0.2\textwidth}{
    \input{images/generation_causal.tikz}}
    \caption{A causal graph for data generation process.} \label{fig:generation_causal}
\end{figure}
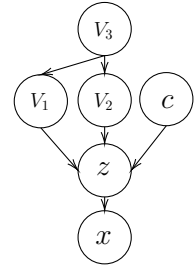

\textbf{Data generation} is assumed to be a generation function that takes ground truth latent representation as the input to generate an observation data/image. 

Formally, we assume that the marginal distribution of sampling ground truth representations $\mathbf{z} \subseteq \mathcal{Z} \in \mathbb{R}^{d_1}$ w.r.t. a \textit{class} is uniform on a unit sphere $\mathbb{S}^{d_1-1}$. An injective generation function $g(.):\mathbb{R}^{d_1}\to \mathbb{R}^d$ takes a ground truth representation and generate an observation sample $\mathbf{x}\in \mathbb{R}^d$, $\mathbf{x}=g(\mathbf{z})$. Different variables, denoted as a set $V=\{V_i\}$ and \textit{class/content} variable $c$, constitute the values of each dimension in the ground truth representation. These variables can be direct causes or confounding factors, $\mathbf{z}=(c,\{V_i\})$. Specific examples include \textit{view angles}, \textit{object size}, \textit{background colors}, etc. A simple causal graph depicts this relationship is shown in Figure \ref{fig:generation_causal}. So the general generation process is as following:
\begin{align} \label{eq:generation2}
    p(\mathbf{z} = (c,V_*)) \sim \frac{1}{|\mathcal{Z}|} && \mathbf{x}=g(\mathbf{z})
\end{align}
To sample a positive example w.r.t the same \textit{class}, we assume the conditional distribution is a von Mises-Fisher (vMF) distribution \cite{fisher1953dispersion}:
\begin{equation} \label{eq:generation}
    p(\tilde{\mathbf{z}}=(c,\tilde{V}_*)|\mathbf{z}=(c,V_*)) = C_p^{-1}e^{\kappa_1 \mathbf{z}^{\top}\tilde{\mathbf{z}}}
\end{equation} 
where $C_p=\int{e^{\kappa_1\mathbf{z}^{\top}\tilde{\mathbf{z}}}d{\tilde{\mathbf{z}}}}$ is a normalization constant and $\kappa$ is a concentration parameter.

\textbf{Representation learning} is a process that a feature encoder, $f(.):\mathbb{R}^d\to \mathbb{R}^{d_2}$, extracts representations from two positive observations $f(\tilde{\mathbf{x}})=f\circ g(\tilde{\mathbf{z}}),f(\mathbf{x})=f\circ g(\mathbf{z})$ and a distribution associated with the encoder $f$ through $h=f\circ g$ is:
\begin{equation} \label{eq:representation}
    q_h(\tilde{\mathbf{z}}|\mathbf{z})=C_q^{-1}(\mathbf{z})e^{\kappa_2 h(\tilde{\mathbf{z}})^{\top}h(\mathbf{z})}
\end{equation} 
with $C_q(\mathbf{z})=\int{e^{\kappa_2 h(\tilde{\mathbf{z}})^{\top}h(\mathbf{z})}d\tilde{\mathbf{z}}}$ be the normalization term.\footnote{The mapping of representations on a hypersphere may be different to Barlow Twins methodology, but as shown in \cite{zbontar2021barlow}, normalize representations on a unit sphere also works under Barlow Twins.} Optimizing any discriminative SSL objective will maximize the similarity between these positive pairs. An example of a well constructed objective is the InfoNCE loss:
\begin{equation} \label{eq:infonce}
    \mathcal{L}_{infoNCE}=\mathbb{E}[-log\frac{e^{f(\tilde{\mathbf{x}})^{\top}f(\mathbf{x})/\tau}}{e^{f(\mathbf{\tilde{x}})^{\top}f(\mathbf{x})/\tau}+\sum_{i=1}^K{e^{f(\mathbf{x}_i^-)^{\top}f(\mathbf{x})/\tau}}}]
\end{equation} 
where $\tilde{\mathbf{x}}$ is a positive example w.r.t. $\mathbf{x}$ in the obsevation space and $\{\mathbf{x}_i^-\}_1^K$ are $K$ samples from distributions of all observations. The global minimum of (\ref{eq:infonce}) is reached when the cosine similarity between positive pairs is maximized and the cosine similarity between all negative pairs is minimized. In the following section, we will show that discriminative SSL including InfoNCE-driven and non InfoNCE-driven (EMA and Siamese with a predictor) follows strict rules of alignment to maximize the similarity.

\subsection{Alignment in discriminative SSL}\label{alignment and regularization}
In this section we will combine theories stated in \cite{Zimmermann2021,Von2021,Wang2020} so that the general factors of successful discirminative SSL can be summarized. Additionally, instead of focusing on just InfoNCE SSL variation and content block-identifiability \cite{hyvarinen2000emergence,le2011learning}, we extend the combined theory to demonstrate that all discriminative SSL benefit from \textit{alignment} between positive representations.

\begin{theorem}
With a data generation process described in \ref{formulation}, all discriminative SSL objectives have an alignment loss function between positive pairs from the network:
\begin{equation}
    \mathcal{L}_{align} = \|f(\mathbf{x})-f(\tilde{\mathbf{x}})\|^2_2
\end{equation}
\end{theorem} \label{thm:alignment}
This is a weaker statement than \textbf{Theorem 4.4} in \cite{Von2021} since we only focus on the alignment term. For analysis on the regularization term on the network entropy, see \ref{appendix:extend}.

\noindent\textbf{Proof}. For InfoNCE-driven SSL (SimCLR and MoCo), as derived in \cite{wang2018deep,Zimmermann2021}, the InfoNCE loss converges to an \textit{alignment} term and a \textit{uniformity} term as the number of negative samples approaches infinity. (See \ref{appendix:extend} for full details).

For EMA-based SSL methods (BYOL), a predictor $p$ is associated with the online network so the SSL objective becomes $\mathcal{L}=\mathop{\mathbb{E}}_{(\mathbf{x},\tilde{\mathbf{x}})}\|p\circ f(\mathbf{x},\theta)-f(\tilde{\mathbf{x}},\xi)\|^2_2$, where $\xi^t=\alpha\xi^{t-1}+(1-\alpha)\theta^t$ is target network parameter and $\theta$ is the online network parameter. Denote $p'=p\circ f$. By adding and subtracting $p'(\tilde{\mathbf{x}},\theta)$ we derive:
\begin{align}
    \begin{split}
        \mathcal{L} = \mathop{\mathbb{E}}_{(\mathbf{x},\tilde{\mathbf{x}})}\|p'(\mathbf{x},\theta)-p'(\tilde{\mathbf{x}},\theta)+p'(\tilde{\mathbf{x}},\theta)-f(\tilde{\mathbf{x}},\xi)\|^2_2 \\
        = \mathop{\mathbb{E}}_{(\mathbf{x},\tilde{\mathbf{x}})}\|p'(\mathbf{x},\theta)-p'(\tilde{\mathbf{x}},\theta)\|^2_2+\mathop{\mathbb{E}}_{\tilde{\mathbf{x}}}\|p'(\tilde{\mathbf{x}},\theta)-f(\tilde{\mathbf{x}},\xi)\|^2_2 \\
        - 2\mathop{\mathbb{E}}_{(\mathbf{x},\tilde{\mathbf{x}})}[(p'(\tilde{\mathbf{x}},\theta)-p'(\mathbf{x},\theta))^{\top}(p'(\tilde{\mathbf{x}},\theta)-f(\tilde{\mathbf{x}},\xi))]
    \end{split} \label{eq:ema}
\end{align}
Since $f(\mathbf{x},\theta)$ and $f(\tilde{\mathbf{x}},\xi)$ maps in the same space $\mathbb{R}^{d_2}$, $p$ can be considered as a bijective linear transformation within $\mathbb{R}^{b_2}$. In fact, the performance difference between a linear predictor and non-linear predictor is subtle. The global minimizer of (\ref{eq:ema}) must align network output w.r.t to $(\mathbf{x},\tilde{\mathbf{x}})$ with the first term in (\ref{eq:ema}) and align outputs from different networks with second term in (\ref{eq:ema}). Hence completes the proof.

For Siamese network with a predictor (SimSiam), a similar approach can refomulate the objective as it is a special case when $f(\tilde{\mathbf{x}},\xi)=f(\tilde{\mathbf{x}},\theta)$ and a \texttt{stop-gradient} is applied on the $f(\tilde{\mathbf{x}},\theta)$. Hence by substuting $\texttt{sg}(f(\tilde{\mathbf{x}},\theta))$
with $f(\tilde{\mathbf{x}},\xi)$ in (\ref{eq:ema}), we complete the proof by:
\begin{align}
    \begin{split}
        \mathcal{L} &=\mathop{\mathbb{E}}_{(\mathbf{x},\tilde{\mathbf{x}})}\|p'(\mathbf{x},\theta)-p'(\tilde{\mathbf{x}},\theta)\|^2_2\\
        &+\mathop{\mathbb{E}}_{\tilde{\mathbf{x}}}\|p'(\tilde{\mathbf{x}},\theta)-\texttt{sg}(f(\tilde{\mathbf{x}},\theta))\|^2_2 \\
        - 2\mathop{\mathbb{E}}_{(\mathbf{x},\tilde{\mathbf{x}})}&[(p'(\tilde{\mathbf{x}},\theta)-p'(\mathbf{x},\theta))^{\top}(p'(\tilde{\mathbf{x}},\theta)-\texttt{sg}(f(\tilde{\mathbf{x}},\theta)))]
        \end{split} \label{eq:siamese}
        \end{align}
Note that the third term in (\ref{eq:ema}) and (\ref{eq:siamese}) can be formulated by the differential entropy of $H(p'(.))$ hence prevent representation collapse.

For Barlow Twins, the diagonal of the cross-correlation matrix $C_{ii}$ is the cosine similarity between positive pairs. Hence completes the proof by:
\begin{align}
    \mathop{\sum}_{i}^{d_2}(1-C_{ii})^2 = \mathop{\sum}_{i}(1-(f(\mathbf{x})^{\top}f(\tilde{\mathbf{x}}))_i/(d_2-1))^2\\
    = \mathop{\sum}_{i}(\|f(\mathbf{x})_i-f(\tilde{\mathbf{x}})_i\|_2^2/(2*(d_2-1))^2
\end{align}

\subsection{Transformation of the ground-truth factors is orthogonal to the applied augmentations}\label{augmentation orthogonal}
By demonstrating that all discriminative SSL have a alignment loss term, the transformation between the ground truth representations and the inferred representations can be derived as in \cite{Zimmermann2021}. But different to \cite{Zimmermann2021}, our derivation of minimization of cross entropy is assumed to be a lower bound since we only include the alignment term and the uniformity term is always positive \cite{cohn2007universally,borodachov2019discrete}. However, with all SSL objectives there are additional terms to maximize the output entropy of the model (some described in \ref{alignment and regularization}). So optimizing SSL objectives as a complete loss function will minimize the cross entropy $\mathbb{E}[H(p(.|\mathbf{z}),q_h(.|\mathbf{z}))]$.

\begin{theorem} \label{thm:match}
    By considering the generation conditional distribution as $p(\tilde{\mathbf{z}}|\mathbf{z}) = C_p^{-1}e^{\kappa_1 \mathbf{z}^{\top}\tilde{\mathbf{z}}}$, the inferred conditional distribution $q_h(\tilde{\mathbf{z}}|\mathbf{z})$ can match $p(\tilde{\mathbf{z}}|\mathbf{z})$ by minimizing $\|h(\mathbf{z})-h(\tilde{\mathbf{z}})\|^2_2$ and $\forall\mathbf{z},\tilde{\mathbf{z}}:\kappa\mathbf{z}^{\top}\tilde{\mathbf{z}}=h(\mathbf{z})^{\top}\tilde{\mathbf{z}}$ with $h=f\circ g$ or $h=p\circ f\circ g$ maps onto a hypersphere with radius $\sqrt{\kappa_1/\kappa_2}$.
\end{theorem}
The proof exactly follows \textbf{Proposition 1} in \cite{Zimmermann2021} just with minor modification on the concentration term $\kappa$ and $h$ so that \textbf{Theorem} \ref{thm:match} can apply to non-InfoNCE SSL. A global minimizer of the alignment term $\|h(\mathbf{z})-h(\tilde{\mathbf{z}})\|^2_2$ will also minimize the cross entropy of between $p(\tilde{\mathbf{z}}|\mathbf{z})$ in (\ref{eq:generation}) and $q_h(\tilde{\mathbf{z}}|\mathbf{z})$ in (\ref{eq:representation}). This indicates that minimizers of SSL objective alignment maintain the dot product. Then we can use \textbf{Proposition 2} in \cite{Zimmermann2021} directly to show that $h$ is an orthogonal linear transformation. 

\begin{theorem} \label{thm:matrix}
    Assume the data generation process (cf. \ref{formulation}), a model parameterized by $h=f\circ g$ or $h=p \circ f \circ g$ ($h:\mathbb{R}^{d_1}\to \mathbb{R}^{d_2}$) that minimizes the alignment term in all discriminative SSL objectives: $\|h(\mathbf{z})-h(\tilde{\mathbf{z}})\|^2_2$ as (\ref{eq:align and uniform}), $h$ is an orthogonal transformation: $h(\tilde{\mathbf{z}})=\mathbf{A}\tilde{\mathbf{z}}$ where $\mathbf{A}$ is an orthogonal matrix.
\end{theorem}

The proof follows that a function $h$ minimizes the alignment will also minimize the cross entropy between the ground truth conditional distribution $p(\tilde{\mathbf{z}}|\mathbf{z})$ and the inferred conditional distribution $q_h(\tilde{\mathbf{z}}|\mathbf{z})$. Therefore if $h$ is isometric w.r.t the dot product as indicated in \textbf{Theorem} \ref{thm:match} then $h(\tilde{\mathbf{z}})=\mathbf{A}\tilde{\mathbf{z}}$ according to \textbf{Proposition 2} in \cite{Zimmermann2021}.

\begin{theorem}
    Assume augmentations applied during the training can be represented as a change in the ground truth representations, i.e. $\tilde{\mathbf{x}}=g(\tilde{\mathbf{z}}(c,\tilde{V*}))$ and change in data variables induces a shift in the ground truth representations $\delta \mathbf{z}=(c,\tilde{V}_*)-(c,V_*)$, then $\delta \mathbf{z}$ is in orthogonal to $\mathbf{A}$ i.e. $\mathbf{A}\delta \mathbf{z}=0$ if a discriminative SSL objective is optimized.
\end{theorem}

\noindent\textbf{Proof}. With augmentations normally applied during SSL such as color distortions, rotations, random cropping, and etc., one can view the alteration as a change of a variable in the data variable (color, view angles, sizes). This change of variable under the generation framework described in \ref{formulation} will result in a change in $V=\{V_i\}$ since $c$ is not changed. Regardless of structure of the causal graph shown in Figure \ref{fig:generation_causal}, the change in $V$ can be reflected in the ground truth representation space as $\tilde{\mathbf{z}}=\mathbf{z}+\delta \mathbf{z}$. A global minimizer of any discriminative SSL objective will minimize the alignment term $h(\mathbf{z})=h(\tilde{\mathbf{z}})=h(\mathbf{z}+\delta \mathbf{z})$. According to \textbf{Theorem} \ref{thm:matrix} we derive:
\begin{align}
    h(\mathbf{z})=\mathbf{A}\mathbf{z}&=\mathbf{A}(\mathbf{z}+\delta \mathbf{z}) \\
    \mathbf{A}\delta \mathbf{z} &= 0
\end{align}
with $\mathbf{A}$ being a linear orthogonal matrix. This indicates that the transformation $A$ is learned to annul the change in data variables or the effect of augmentations applied. 

\subsection{Reason and solutions for unstable change in data variables}\label{solution}
In section \ref{augmentation orthogonal} we show that generalized representation is robust to augmentations since $\mathbf{A}$, the linear transformation between the ground truth representation and inferred representations, is orthogonal to augmentation applied during the training. And if a change in the data variable reflects a shift, $\delta \mathbf{z}$, in the ground truth representations and the newly inferred representation is stable, meaning $\delta \mathbf{z}$ will be absorbed by the transformation matrix $\mathbf{A}$, then the shift in the ground truth representation corresponds to an augmentation that is applied during the training. However, when $\delta \mathbf{z}$ appears in the range of $\mathbf{A}$ and $\mathbf{A}\delta\mathbf{z}\neq{0}$, the resultant inferred representation can be unstable and lead to performance drop on downstream models (denote as $D(\mathbf{z})$). We quantify this deterioration on $D$ by:
\begin{equation}
    m(D(h(\mathbf{z}))_{stable})-m(D(h(\tilde{\mathbf{z}})_{unstable})
\end{equation}
where $m(.)$ is a metric on an outcome of the downstream task. In example of $D$ being a linear classifier, $m(.)$ can be the probability of predicting the target class (\textbf{prediction score}) or the proportion of correct predictions (\textbf{accuracy}). In order to overcome this deterioration, we propose two methods, namely \textbf{Robust Dimensions} and \textbf{Stable Inference Mapping}: 
\begin{enumerate}
    \item \textbf{Robust Dimensions}: Under a stable condition, $D(f(\mathbf{x}))\footnote{Note that in general $f$ contains a projector. However we exclude the notion of the projector to simplify the problem and \cite{chen2020simple,he2020momentum,grill2020bootstrap} show that a projector is not necessary in SSL.}=D(\mathbf{A}\mathbf{z})$, each dimension in the inferred representation $f(\mathbf{x})$ is a linear combination of dimensions of the ground truth representation. The dimensions contributing most to $D()$ should be robust to unstable shift $\delta \mathbf{z}_{unstable}$. In other words, most robust dimensions of $f(\mathbf{x})$ should be also robust in $f(\tilde{\mathbf{x}})$ where $\tilde{\mathbf{x}}=g(\mathbf{z}+\delta\mathbf{z}_{unstable})$ as some dimensions of $\mathbf{A}\delta \mathbf{z}_{unstable}$ will be zero. Hence identifying most important dimensions in $f(\mathbf{x})_{stable}$ and pass through the same dimensions of $f(\mathbf{\tilde{x}})_{unstable}$ should alleviate the deterioration by making $m(D(h(\mathbf{z}))_{stable})_{\{dim\}}=m(D(h(\tilde{\mathbf{z}})_{unstable})_{\{dim\}}$ where $\{dim\}$ is a set of most robust dimensions. In example of $D$ being a linear classifier, the contribution of each dimension can be calculated by $W^{\top}_cf(\mathbf{x})$ where $W^{\top}_c$ is the Jacobian of the linear classifier w.r.t target class $c$.
    \item \textbf{Stable Inference Mapping}: Since $f(\mathbf{x})_{stable}-f(\tilde{\mathbf{x}})_{unstable}=-\mathbf{A}\delta\mathbf{z}$, we can learn another linear transformation $\mathbf{F}$ to absorb $\delta\mathbf{z}$. Especially, we want to learn $\mathbf{F}f(\tilde{\mathbf{z}})_{unstable}=\mathbf{FAz}+\mathbf{FA}\delta\mathbf{z}$ such that the additional $\mathbf{F}$ will not only set $\mathbf{A}\delta{z}$ to 0, but also make stable representation more robust to the unstable shift by assuring $\mathbf{FA}$ orthogonal to $\delta \mathbf{z}_{unstable}$ in addition to the augmentations applied during training. Formally, we model a linear layer $l(f(\tilde{\mathbf{x}})_{unstable}) = f(\mathbf{x})_{stable}-f(\mathbf{\tilde{x}})_{unstable}$, hence during the inference $f(\mathbf{x})+l(f(\mathbf{x}))$ is used for downstream task.
\end{enumerate}

\textbf{Relation to Causal Inference} Since we only have observations in the image space, we consider the augmentations or changes of a data variable as \textit{interventions} on the ground truth representations: $\tilde{\mathbf{x}}=g(\mathbf{\tilde{z}}|(c,do(V_i=v_i))$. With an access to ground truth representation, we can evaluate treatment-control effect i.e. $Pr(f(g(\mathbf{\tilde{z}}|(c,do(V_i=v_{stable})))-Pr(f(g(\mathbf{\tilde{z}}|(c,do(V_i=v_{unstable})))$. Without any access to the ground truth representations, we can evaluate the average treatment effect \cite{holland1986statistics} by $\mathbb{E}[D(f(\mathbf{x})_{stable})-D(f(\mathbf{\tilde{x}})_{unstable})]$ via synthesizing manual data samples $\tilde{\mathbf{x}_i}=g(\mathbf{\tilde{z}}|(c,do(V_i=v_i))$.

\section{Experiments and Discussions}
In this section we evaluate our solutions to unstable shift in data variable on two datasets: Causal3DIdent and ImageNet pretrained SSL and corresponding linear classifiers as the downstream task.

\subsection{Causal3DIdent}
\begin{figure}[htbp]
    \centering
    \resizebox{0.3\textwidth}{!}{
    \input{images/causal3dident.tikz}
    }
    \caption{Causal graph for $V$ in Causal3DIdent.}
    \label{fig:causal3d}
\end{figure}
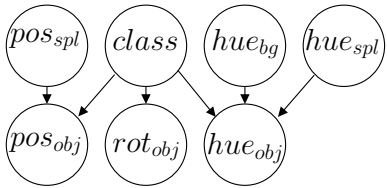

\begin{figure*}[htbp]
    \centering
    \begin{subfigure}[b]{0.48\textwidth}
          \centering
          \resizebox{\linewidth}{!}{\includegraphics{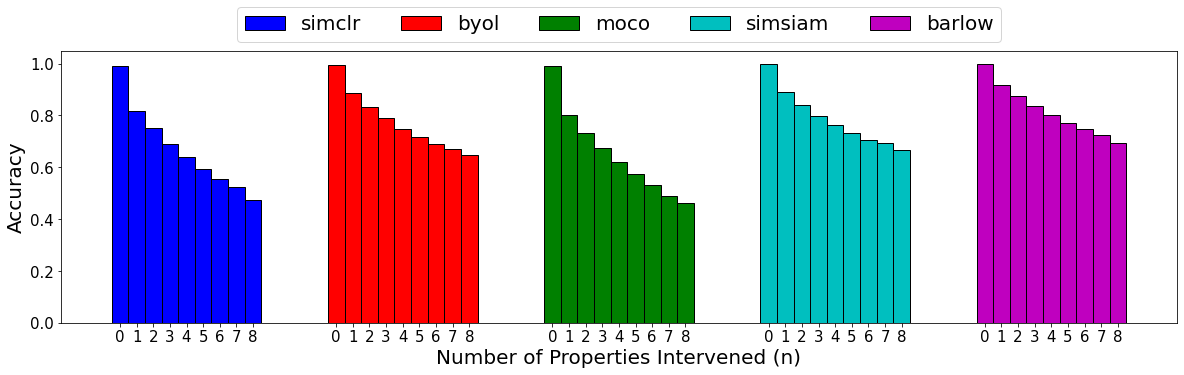}}
          \caption{Deterioration in Accuracy}
          \label{fig:causal_parents_acc}
     \end{subfigure}
     \begin{subfigure}[b]{0.48\textwidth}
          \centering
          \resizebox{\linewidth}{!}{\includegraphics{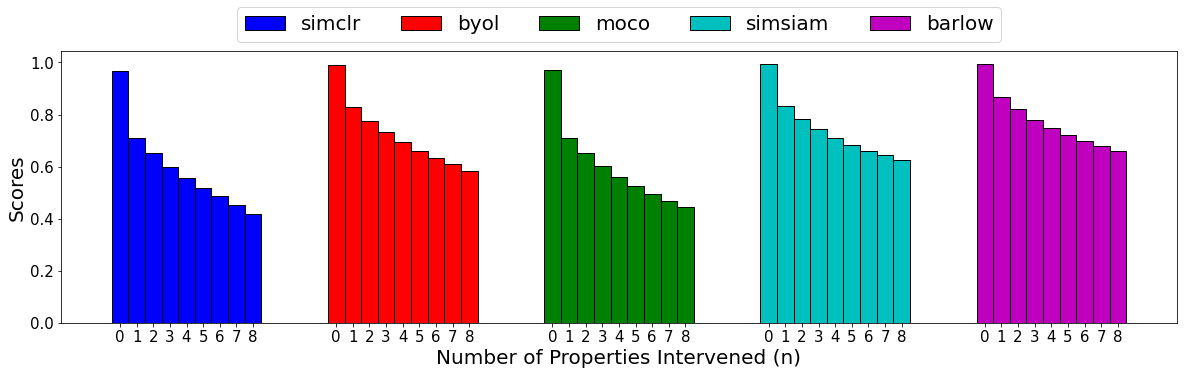}}  
          \caption{Deterioration in Score}
          \label{fig:causal_parents_score}
     \end{subfigure}
    \caption{Deterioration of unstable changing in data variables on all SSL. The more unexpected changes occur in the data variables, the more severe the deterioration.}
    \label{fig:causal_parents}
\end{figure*}
\cite{Zimmermann2021} develops the dataset that 7 objects in the dataset that each $224\times224$ image is associated with an object class and a 10-dimensional latent representation. These 10 dimensions correspond to 10 data variables. A causal graph imposed on these variables is shown in Figure \ref{fig:causal3d}. As the dependency shown in Figure \ref{fig:causal3d}, an object is placed at $pos_{obj}=(x,y,z)$ with $rot_{obj}=(\phi,\theta,\psi)$ and $hue_obj$ under a spot light at an angle $pos_{spl}$ with color $hue_{spl}$ on a background with color $hue_{bg}$. Detailed information in \ref{appendix:causal3d}.

To simulate unstable change of variable that is not accessible during the training, a range of some data variables is hold out and hence this portion of data is treated as potential unstable samples that causes the performance drop on the linear classifier. Since all data variables are truncated in range [-1,1], we hold out the edge value(s) to further portray 'unexpected' values during the inference time. We select 8 dimensions to intervene, namely: $z$ in object position\footnote{Only $z$ is altered because most of deep vision models are translation invariant.}, all 3 object rotation angles, spot light position, and all 3 hue variables. Both training and testing data are sampled and for detailed sampling procedures refer to \ref{appendix:causal3d}. 

\noindent\textbf{SSL Experiment Setup} ResNet18\cite{he2016deep} is the backbone of the feature encoder $f$. Same augmentations in \cite{chen2020simple} are applied during training. An Adam optimizer with a learning rate at 0.0001 and a weight decay at 0.00001 is optimized for all discriminative SSL. Hyperparameters for each SSL is presented in \ref{appendix:causal3d}. The dimension of the inferred representation space is set to 128. The network is trained for 20 epochs on intervened data. Then a linear classifier is trained on the frozen representations of the network with a same optimizer for 10 epochs.

To illustrate the simulation results in the deterioration of the downstream performance, for each testing data that has the same data variable distribution as the training data, we search in the ground truth representation space for 5 nearest representations in the hold-out test distribution when $n$ dimension of the testing data is shifted to a random value in the hold-out range of corresponding data variable. Within the corresponding 5 images representing a shift in $n$ dimensions that is not seen during the training, we record the lowest performance to fulfill the deterioration scenario. In Figure \ref{fig:causal_parents} the score and accuracy are averaged over all combinations of $\binom{8}{n}$. When $n=0$, this indicates the performance of SSL on the testing data with the same data variable distribution as the training data. The performance is exceptional since hold-out values of some data variables may be covered by the augmentations applied ($z$ covered by random cop, and hue colors covered by color distortions), $\mathbf{A}$ will be orthogonal to changes in these data variables. In fact in \ref{appendix:causal3d} we show that the performance difference between seen distribution and unseen distribution in testing data is comparatively small. As illustrated in Figure \ref{fig:causal_parents}, even when only one variable is changed to a unseen value, there is a large drop in both accuracy and prediction score ($~20\%$ in accuracy and $~30\%$ in prediction score for SimCLR). With more tangled changes in data variables, the unstable representations results in poorer downstream performance. Since the selection of the data variables may be too complex due to the dependency, we also validate the same issue on selecting only children nodes in Figure \ref{fig:causal3d}. And also we visualize the latent shift between stable and unstable examples. See \ref{appendix:causal3d} for more results.

\begin{figure}[htbp]
    \centering
    \begin{subfigure}[b]{0.5\textwidth}
          \centering
          \resizebox{\linewidth}{!}{\includegraphics{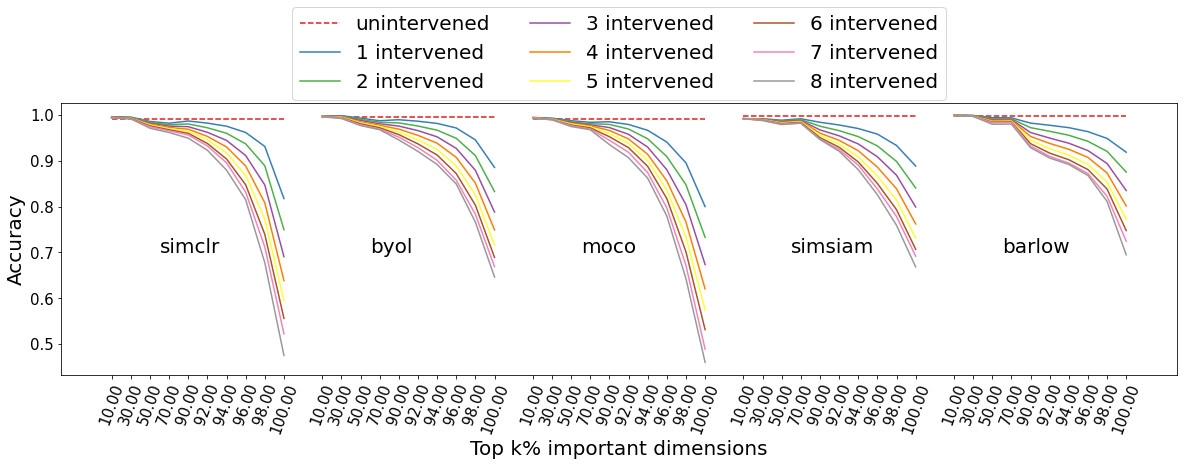}}
          \caption{Improvement on Accuracy}
          \label{fig:s1_parents_acc}
     \end{subfigure}
     \begin{subfigure}[b]{0.5\textwidth}
          \centering
          \resizebox{\linewidth}{!}{\includegraphics{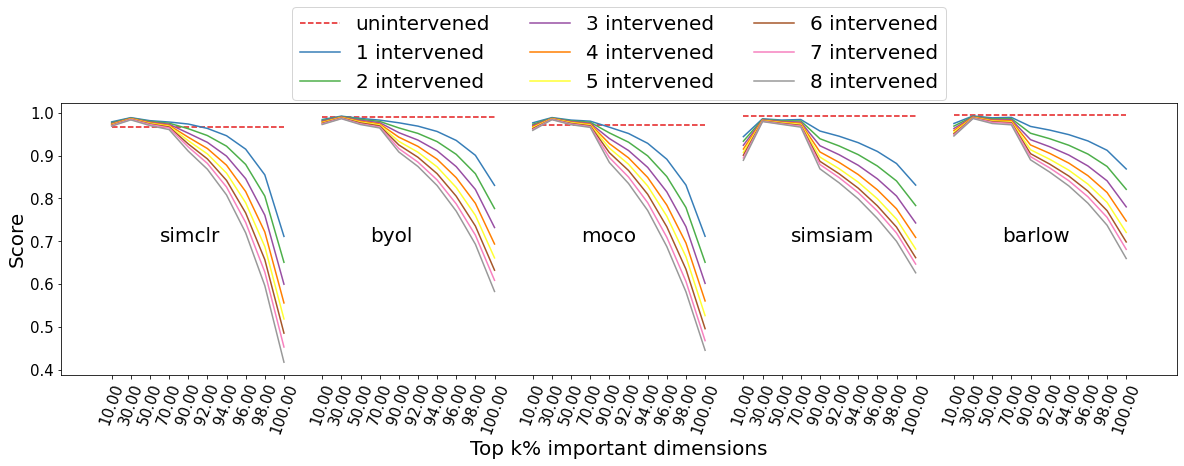}}  
          \caption{Improvement on Score}
          \label{fig:s1_parents_score}
     \end{subfigure}
    \caption{By identifying and passing top $k$ \% important dimensions, the deterioration in both accuracy and prediction score is significantly alleviated.}
    \label{fig:s1_parents}
\end{figure}

\noindent\textbf{Robust Dimensions} For each pair of testing data $\mathbf{x}_{stable}$ with seen data variable distribution and the selected data $\mathbf{x}_{unstable}$ among the 5 nearest neighbours when changing $n$ dimensions to unseen distribution values, we apply the Jacobian of the linear classifier w.r.t the target class $W^{\top}_c$ on the stable representation $f(\mathbf{x})_{stable}$ to identify the top $k$ most important dimensions and pass the same dimensions of $f(\tilde{\mathbf{x}})_{unstable}$ to $D$ to evaluate the performance. As shown in Figure \ref{fig:s1_parents}, the accuracy only deteriorate slightly when top 90\%
most important features of $f(\mathbf{\tilde{x}})_{unstable}$ are selected for the downstream task. This is true even when all 8 variables are shifted. This suggests that $\mathbf{A}$ is orthogonal to changes in ground truth representation $\mathbf{z}$ in most dimensions. This high percentage of dimensions may be due to $f$ optimizes the SSL objectives to a high level and the augmentations applied covers some of the hold-out variable values. Interestingly, there are cases where passing the top $k$\% (around 40\%) important dimensions results in higher performance than stable representations (unitervened in Figure \ref{fig:s1_parents}). However, as expected, including more less important dimensions where the unseen shift in the data variable results in non-zero adjustment ($\mathbf{A}\delta \mathbf{z}\neq{0}$) initiates the deterioration phenomenon we observe.

\noindent\textbf{Stable Inference Mapping} As shown in \ref{solution}, a linear transformation $\mathbf{F}:\mathbb{R}^{d_2}\to\mathbb{R}^{d_2}$ is trained to cancel the effect $\mathbf{A}\delta \mathbf{z}$ and also further improve the robustness of transformation on the ground truth representation $\mathbf{A}\mathbf{z}$. In this experiment, we match each training data with a random member in the 5 nearest representations when only \textbf{one} dimension is changed to a unstable value. %($\mathbf{x},\mathbf{\tilde{x}}$). 
A linear layer is trained with the same optimizer for 10 epochs on the training pairs. In Table \ref{tbl:f}, all accuracy for unstable examples increases significantly (except SimSiam) after learning $\mathbf{F}$. However, since the $f$ is very close to the global minimizer of the alignment term, the improvement on stable examples cannot be observed. 

\begin{table}[h]
\centering
\begin{tabular}{|l|ll|ll|}
\hline
        & \multicolumn{2}{c|}{$\mathbf{x}_{stable}$}       & \multicolumn{2}{c|}{$\mathbf{\tilde{x}}_{unstable}$}     \\ \hline
        & \multicolumn{1}{c|}{w/o $\mathbf{F}$} & w/ $\mathbf{F}$ & \multicolumn{1}{l|}{w/o $\mathbf{F}$} & w/ $\mathbf{F}$ \\ \hline
SimCLR  & \multicolumn{1}{c|}{0.996}      &   0.998   & \multicolumn{1}{c|}{0.833}      &  0.889    \\ \hline
MoCo    & \multicolumn{1}{c|}{0.992}      &    0.992  & \multicolumn{1}{c|}{0.800}      &   0.849   \\ \hline
BYOL    & \multicolumn{1}{c|}{0.996}      &    0.998  & \multicolumn{1}{c|}{0.886}      &   0.920   \\ \hline
SimSiam & \multicolumn{1}{c|}{0.998}      &    0.999  & \multicolumn{1}{c|}{0.919}      &    0.928  \\ \hline
Barlow Twins & \multicolumn{1}{c|}{0.991}      &    0.995  & \multicolumn{1}{c|}{0.818}      &  0.862    \\ \hline
\end{tabular}
\caption{The effect of $\mathbf{F}$ on both stable and unstable samples.The accuracy is average over 3 random seeds.} \label{tbl:f}
\end{table}

\subsection{ImageNet}
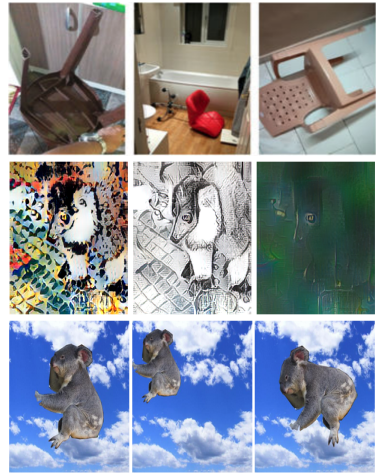
\begin{figure}
    \centering
    \resizebox{0.3\textwidth}{!}{
    \input{images/datasets.tikz}
    }
    \caption{Examples of ObjectNet dataset ($1^{st}$ row), Stylized ImageNet dataset ($2^{nd}$ row), Synthetic dataset ($3^{rd}$ row).}
    \label{fig:datasets}
\end{figure}

To validate our findings on a larger scale and more realistic settings, we apply proposed solutions on ImageNet\cite{deng2009imagenet} with various altered and synthetic datasets as unstable shift in the data variable. See Figure \ref{fig:datasets} and \ref{appendix:imagenet} for more information on \textbf{ObjectNet, Stylized-ImageNet, Synthetic dataset}. 

\textbf{ObjectNet}\cite{barbu2019objectnet} is a collection of objects that are intentionally placed at an unusual view angle and backgrounds, so that the bias learned by the model with usual data variables is more prominent when testing on ObjectNet. With focusing on 113 classes overlapping with ImageNet classes, we evaluate our second proposed solution explicitly on ObjectNet since the shift presented in ObjectNet is very unstable and it is very challenging to overcome the negative effect of the shift.

\textbf{Stylized ImageNet}\cite{geirhos2018} change the style of original ImageNet image to a random artistic style. With this drastic shift in the data variable, the analysis of the first solution is more insightful on the robustness of dimensions of $\mathbf{A}$ on a very different shift in data variable. 

\textbf{Synthetic Data} follows synthetic procedure in \cite{djolonga2020robustness} where object is masked on a background at a location with a rotation angle. We explore the benefit of this synthesized dataset at three modes: \textit{background, location, rotation} where the target variable is randomly sampled with the other two variables fixed. Additionally, we explore \textit{texture} as an independent variable by masking 'texturized' objects in \cite{Geirhos2019}. We set total number of updating steps per epoch as 100 with batch size 256. This means we select total number of 512000 synthesized images every epoch.

\noindent\textbf{Experiment Setup} ResNet 50 pretrained on ImageNet and a linear classifier fintuned via SimCLR, BYOL, and SimSiam are tested with both proposed solutions. For Stable Inference Mapping, a linear layer is optimized with an Adam optimizer for 10 epochs on the synthetic dataset.

\begin{figure}[htbp]
    \centering
    \begin{subfigure}[b]{0.48\textwidth}
          \centering
          \resizebox{\linewidth}{!}{\includegraphics{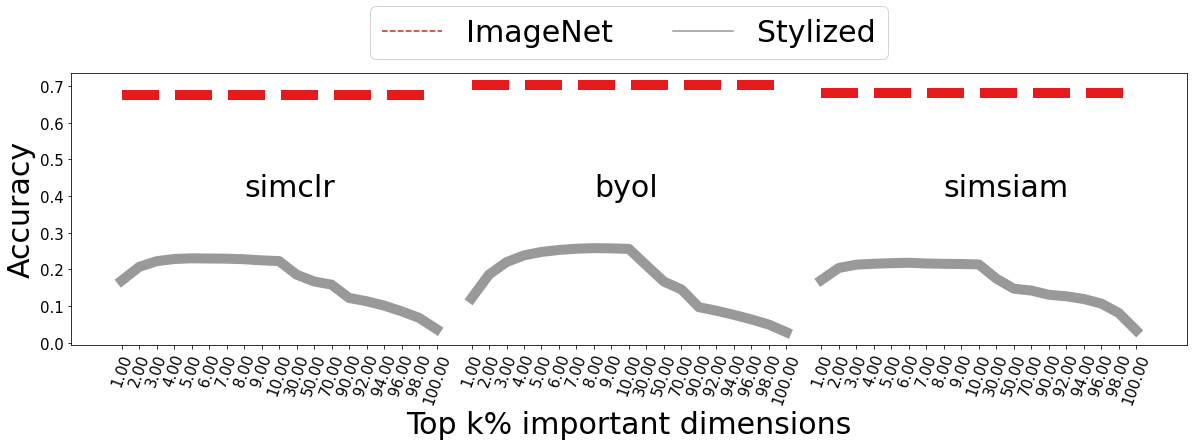}}
          \caption{Improvement on Accuracy}
          \label{fig:s1_style_acc}
     \end{subfigure}
     \begin{subfigure}[b]{0.48\textwidth}
          \centering
          \resizebox{\linewidth}{!}{\includegraphics{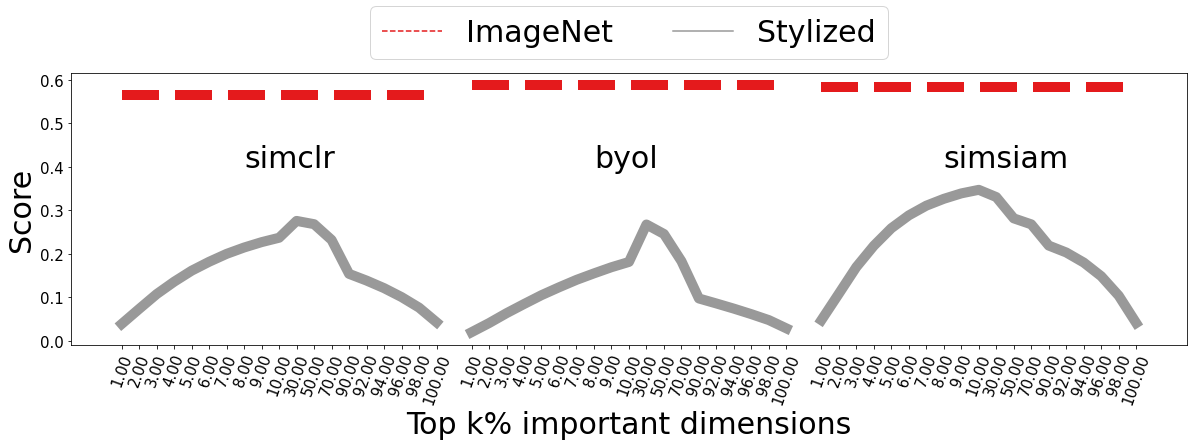}}  
          \caption{Improvement on Score}
          \label{fig:s1_style_score}
     \end{subfigure}
    \caption{By identifying and passing top $k$ \% important dimensions, the deterioration in both accuracy and prediction score is alleviated.}
    \label{fig:s1_style}
\end{figure}

\noindent\textbf{Robust Dimensions} For each of ImageNet validation data sample, we stylized the image to a random artistic fashion. We observe the dramatic performance difference between the ImageNet stable images and Stylized unstable ImageNet images $(\mathbf{x}_{stable},\mathbf{x}_{unstable})$. The result is shown in Figure \ref{fig:s1_style}. For SimCLR, passing the top 10\% important dimensions can close a small performance gap between stable representations and unstable representations. Nonetheless, all SSL seem to be sensible to the strong style change as they shorten the difference between ImageNet to an modest extent. This is as expected since the Stylized ImageNet changes multiple variables to an extreme value.

\noindent\textbf{Stable Inference Mapping} As described in \textbf{Synthetic Data}, we explore the benefit of our second proposed method to evaluate on a dataset that most of samples are unstable according to learned $\mathbf{A}$. At each training step, with other variables randomly fixed, 10 images with random target variable values are generated. The pair with maximum $m(D(\mathbf{x}))-m(D(\tilde{\mathbf{x}}))$ is selected to train the linear transformation. In Table \ref{tbl:s2_object}, inferring $\mathbf{F}$ via controlling location produces least improvement. This is expected since the network is robust to translation by design. While background, rotation, and texture improves the performance considerably with the consideration on the training time. However, in \ref{appendix:imagenet} we show that training the model longer using \textbf{Stable Inference Mapping} yield less favorable results since the improvement is less significant and starts to saturate at around 30 epochs.

\begin{table}[htbp]
\centering
\begin{tabular}{|c|c|c|c|}
\hline
           & SimCLR & BYOL  & SimSiam \\ \hline
w/o F      & 10.12  & 14.04 & 11.15   \\ \hline
background & 12.34  & 15.43 & 12.45   \\ \hline
location   & 10.35  & 14.13 & 11.17   \\ \hline
rotation   & 12.04  & 15.88 & 12.38     \\ \hline
texture    & 12.79  & 15.06 & 13.11       \\ \hline
\end{tabular} 
\caption{Comparison of inferring different $\mathbf{F}$ on ObjectNet with a target data variable random trials.}
\label{tbl:s2_object}
\end{table}

\section{Limitations}
Though we identify the root cause of unstable inference for descriminative SSL by constructing a causal framework inspired by the prior work, the proposed solutions are constrained and limited to be applied on realistic applications. \textbf{Robust Dimensions} involves establishing a correspondence between stable and unstable instances on a one-to-one basis, enabling the identification of dimensions contributing to stability. On the other hand, \textbf{Stable Inference Mapping} necessitates a collection of unstable instances with a specific alteration in a particular group of data variables. Within the Causal3dIdent dataset, both solutions can be assessed using the same unstable instances. In more realistic datasets, achieving a one-to-one correspondence is feasible, and manipulation of one group of data variables can be accomplished using synthetic data. However, any assessments with involving artificially generated images might introduce some level of uncertainty. In a realistic setup, since training samples are not directly observable during the inference stage, simple interventions on inference samples may not effectively separate the unstable variables from the stable ones. Consequently, the potential benefits of the proposed solutions in realistic datasets are undermined.

\section{Conclusions}
In conclusion, this paper has proposed a novel approach to address the issue of unstable behavior during the inference stage in SSL methods. By building on the previous theories of successful InfoNCE-facilitated contrastive SSL and extending it to recent SSL methods, we have demonstrated that a change in the data factor can result in a shift in the inferred representation, leading to a decline in downstream performance. We have proposed learning targeted transformations that regularize the violating shift and restore performance on the unseen data shift. Our experiments on both controlled and realistic datasets have shown the efficacy of our proposed solutions. These contributions provide a better understanding of SSL methods and offer a promising solution to the problem of unstable behavior during the inference stage. We hope that our work will inspire further research in this area and lead to improved SSL methods that are more robust to changes in data factors.

\section{Acknowledgements}
The work was funded in part by NSF CNS-2112562 and IIS-2140247.

{\small
\bibliographystyle{ieee_fullname}
\bibliography{main.bib}
}

\clearpage
\newpage
\appendix
\input{appendix}

\end{document}

%% file: images/problem.tikz
\tikzset{every picture/.style={line width=0.75pt}} %set default line width to 0.75pt        

\begin{tikzpicture}[x=0.75pt,y=0.75pt,yscale=-1,xscale=1]
%uncomment if require: \path (0,520); %set diagram left start at 0, and has height of 520

%Rounded Rect [id:dp00655204637063278] 
\draw  [fill={rgb, 255:red, 74; green, 144; blue, 226 }  ,fill opacity=0.4 ] (135,44) .. controls (135,19.7) and (154.7,0) .. (179,0) -- (531,0) .. controls (555.3,0) and (575,19.7) .. (575,44) -- (575,176) .. controls (575,200.3) and (555.3,220) .. (531,220) -- (179,220) .. controls (154.7,220) and (135,200.3) .. (135,176) -- cycle ;
%Rounded Rect [id:dp29349429630559043] 
\draw  [fill={rgb, 255:red, 184; green, 233; blue, 134 }  ,fill opacity=0.5 ] (135,274) .. controls (135,249.7) and (154.7,230) .. (179,230) -- (531,230) .. controls (555.3,230) and (575,249.7) .. (575,274) -- (575,406) .. controls (575,430.3) and (555.3,450) .. (531,450) -- (179,450) .. controls (154.7,450) and (135,430.3) .. (135,406) -- cycle ;
%Image [id:dp5884031119470547] 
\draw (175,125) node  {\includegraphics[width=37.5pt,height=52.5pt]{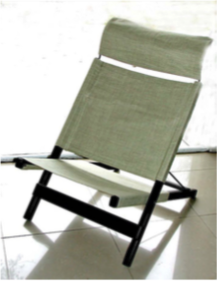}};
%Image [id:dp45397062537264765] 
\draw (245,175) node  {\includegraphics[width=37.5pt,height=52.5pt]{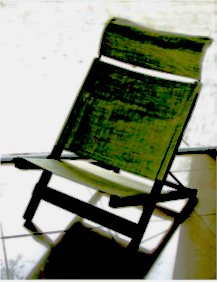}};
%Image [id:dp699829089286047] 
\draw (245,75) node  {\includegraphics[width=37.5pt,height=52.5pt]{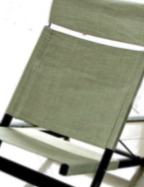}};
%Rounded Rect [id:dp1832098524942214] 
\draw   (310,48) .. controls (310,43.58) and (313.58,40) .. (318,40) -- (342,40) .. controls (346.42,40) and (350,43.58) .. (350,48) -- (350,202) .. controls (350,206.42) and (346.42,210) .. (342,210) -- (318,210) .. controls (313.58,210) and (310,206.42) .. (310,202) -- cycle ;
%Shape: Rectangle [id:dp551229909053323] 
\draw  [color={rgb, 255:red, 0; green, 0; blue, 0 }  ,draw opacity=1 ][fill={rgb, 255:red, 74; green, 144; blue, 226 }  ,fill opacity=1 ] (390,45) -- (410,45) -- (410,105) -- (390,105) -- cycle ;
%Shape: Rectangle [id:dp017697759058038054] 
\draw  [color={rgb, 255:red, 0; green, 0; blue, 0 }  ,draw opacity=1 ][fill={rgb, 255:red, 74; green, 144; blue, 226 }  ,fill opacity=1 ] (390,145) -- (410,145) -- (410,205) -- (390,205) -- cycle ;
%Straight Lines [id:da8900718076325926] 
\draw [color={rgb, 255:red, 208; green, 2; blue, 27 }  ,draw opacity=1 ]   (175,90) -- (175,75) -- (218,75) ;
\draw [shift={(220,75)}, rotate = 180] [color={rgb, 255:red, 208; green, 2; blue, 27 }  ,draw opacity=1 ][line width=0.75]    (10.93,-3.29) .. controls (6.95,-1.4) and (3.31,-0.3) .. (0,0) .. controls (3.31,0.3) and (6.95,1.4) .. (10.93,3.29)   ;
%Straight Lines [id:da1117672575721218] 
\draw [color={rgb, 255:red, 208; green, 2; blue, 27 }  ,draw opacity=1 ]   (175,160) -- (175,175) -- (218,175) ;
\draw [shift={(220,175)}, rotate = 180] [color={rgb, 255:red, 208; green, 2; blue, 27 }  ,draw opacity=1 ][line width=0.75]    (10.93,-3.29) .. controls (6.95,-1.4) and (3.31,-0.3) .. (0,0) .. controls (3.31,0.3) and (6.95,1.4) .. (10.93,3.29)   ;
%Straight Lines [id:da8231080143804992] 
\draw    (270,75) -- (308,75) ;
\draw [shift={(310,75)}, rotate = 180] [color={rgb, 255:red, 0; green, 0; blue, 0 }  ][line width=0.75]    (10.93,-3.29) .. controls (6.95,-1.4) and (3.31,-0.3) .. (0,0) .. controls (3.31,0.3) and (6.95,1.4) .. (10.93,3.29)   ;
%Straight Lines [id:da10577718534583092] 
\draw    (270,175) -- (308,175) ;
\draw [shift={(310,175)}, rotate = 180] [color={rgb, 255:red, 0; green, 0; blue, 0 }  ][line width=0.75]    (10.93,-3.29) .. controls (6.95,-1.4) and (3.31,-0.3) .. (0,0) .. controls (3.31,0.3) and (6.95,1.4) .. (10.93,3.29)   ;
%Straight Lines [id:da868152043225008] 
\draw    (350,75) -- (388,75) ;
\draw [shift={(390,75)}, rotate = 180] [color={rgb, 255:red, 0; green, 0; blue, 0 }  ][line width=0.75]    (10.93,-3.29) .. controls (6.95,-1.4) and (3.31,-0.3) .. (0,0) .. controls (3.31,0.3) and (6.95,1.4) .. (10.93,3.29)   ;
%Straight Lines [id:da3421615256316519] 
\draw    (350,175) -- (388,175) ;
\draw [shift={(390,175)}, rotate = 180] [color={rgb, 255:red, 0; green, 0; blue, 0 }  ][line width=0.75]    (10.93,-3.29) .. controls (6.95,-1.4) and (3.31,-0.3) .. (0,0) .. controls (3.31,0.3) and (6.95,1.4) .. (10.93,3.29)   ;
%Image [id:dp8933794798921093] 
\draw (175,285) node  {\includegraphics[width=37.5pt,height=52.5pt]{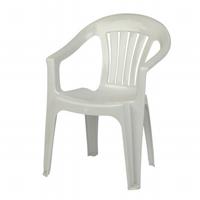}};
%Rounded Rect [id:dp25428560003499245] 
\draw   (230,248) .. controls (230,243.58) and (233.58,240) .. (238,240) -- (262,240) .. controls (266.42,240) and (270,243.58) .. (270,248) -- (270,402) .. controls (270,406.42) and (266.42,410) .. (262,410) -- (238,410) .. controls (233.58,410) and (230,406.42) .. (230,402) -- cycle ;
%Image [id:dp04689664295697882] 
\draw (175,365) node  {\includegraphics[width=37.5pt,height=52.5pt]{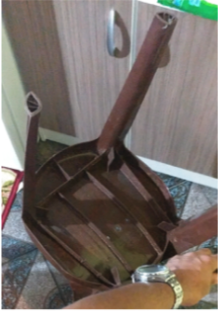}};
%Shape: Trapezoid [id:dp8472086243379484] 
\draw   (349,245) -- (414,264.5) -- (414,390.5) -- (349,410) -- cycle ;
%Shape: Rectangle [id:dp27637710076673216] 
\draw  [color={rgb, 255:red, 0; green, 0; blue, 0 }  ,draw opacity=1 ][fill={rgb, 255:red, 74; green, 144; blue, 226 }  ,fill opacity=1 ] (297,255) -- (317,255) -- (317,315) -- (297,315) -- cycle ;
%Shape: Rectangle [id:dp21437628878751736] 
\draw  [color={rgb, 255:red, 0; green, 0; blue, 0 }  ,draw opacity=1 ][fill={rgb, 255:red, 208; green, 2; blue, 27 }  ,fill opacity=1 ] (297,340) -- (317,340) -- (317,400) -- (297,400) -- cycle ;
%Shape: Right Angle [id:dp2544932842837748] 
\draw   (563,399.5) -- (469.29,399.83) -- (468.88,280.62) ;
%Shape: Rectangle [id:dp22565950897619214] 
\draw  [color={rgb, 255:red, 0; green, 0; blue, 0 }  ,draw opacity=1 ][fill={rgb, 255:red, 74; green, 144; blue, 226 }  ,fill opacity=1 ] (487,298) -- (502,298) -- (502,398) -- (487,398) -- cycle ;
%Shape: Rectangle [id:dp886908320013648] 
\draw  [color={rgb, 255:red, 0; green, 0; blue, 0 }  ,draw opacity=1 ][fill={rgb, 255:red, 208; green, 2; blue, 27 }  ,fill opacity=1 ] (533,348) -- (548,348) -- (548,398) -- (533,398) -- cycle ;
%Right Arrow [id:dp6222009358056806] 
\draw  [fill={rgb, 255:red, 126; green, 211; blue, 33 }  ,fill opacity=1 ] (424,320) -- (445,320) -- (445,310) -- (459,330) -- (445,350) -- (445,340) -- (424,340) -- cycle ;
%Straight Lines [id:da6088590528827962] 
\draw    (200,285) -- (228,285) ;
\draw [shift={(230,285)}, rotate = 180] [color={rgb, 255:red, 0; green, 0; blue, 0 }  ][line width=0.75]    (10.93,-3.29) .. controls (6.95,-1.4) and (3.31,-0.3) .. (0,0) .. controls (3.31,0.3) and (6.95,1.4) .. (10.93,3.29)   ;
%Straight Lines [id:da4182391494065647] 
\draw    (200,365) -- (228,365) ;
\draw [shift={(230,365)}, rotate = 180] [color={rgb, 255:red, 0; green, 0; blue, 0 }  ][line width=0.75]    (10.93,-3.29) .. controls (6.95,-1.4) and (3.31,-0.3) .. (0,0) .. controls (3.31,0.3) and (6.95,1.4) .. (10.93,3.29)   ;
%Straight Lines [id:da010454008208369725] 
\draw    (270,285) -- (295,285) ;
\draw [shift={(297,285)}, rotate = 180] [color={rgb, 255:red, 0; green, 0; blue, 0 }  ][line width=0.75]    (10.93,-3.29) .. controls (6.95,-1.4) and (3.31,-0.3) .. (0,0) .. controls (3.31,0.3) and (6.95,1.4) .. (10.93,3.29)   ;
%Straight Lines [id:da6828976804427906] 
\draw    (270,365) -- (295,365) ;
\draw [shift={(297,365)}, rotate = 180] [color={rgb, 255:red, 0; green, 0; blue, 0 }  ][line width=0.75]    (10.93,-3.29) .. controls (6.95,-1.4) and (3.31,-0.3) .. (0,0) .. controls (3.31,0.3) and (6.95,1.4) .. (10.93,3.29)   ;
%Straight Lines [id:da14775638545841407] 
\draw    (320,285) -- (347,285) ;
\draw [shift={(349,285)}, rotate = 180] [color={rgb, 255:red, 0; green, 0; blue, 0 }  ][line width=0.75]    (10.93,-3.29) .. controls (6.95,-1.4) and (3.31,-0.3) .. (0,0) .. controls (3.31,0.3) and (6.95,1.4) .. (10.93,3.29)   ;
%Straight Lines [id:da6968789754416957] 
\draw    (320,365) -- (347,365) ;
\draw [shift={(349,365)}, rotate = 180] [color={rgb, 255:red, 0; green, 0; blue, 0 }  ][line width=0.75]    (10.93,-3.29) .. controls (6.95,-1.4) and (3.31,-0.3) .. (0,0) .. controls (3.31,0.3) and (6.95,1.4) .. (10.93,3.29)   ;

% Text Node
\draw (315,150) node [anchor=north west][inner sep=0.75pt]  [font=\huge,rotate=-270] [align=left] {$\displaystyle f(x)$};
% Text Node
\draw    (438,92.75) -- (561,92.75) -- (561,152.75) -- (438,152.75) -- cycle  ;
\draw (499.5,122.75) node   [align=left] {\begin{minipage}[lt]{80.92pt}\setlength\topsep{0pt}
\begin{center}
{\Large $\displaystyle Align\left(\mathbf{z} ,\mathbf{\tilde{z}}\right)$}
\end{center}

\end{minipage}};
% Text Node
\draw (302,3) node [anchor=north west][inner sep=0.75pt]  [font=\huge] [align=left] {Training};
% Text Node
\draw (417,70) node [anchor=north west][inner sep=0.75pt]  [font=\huge]  {$\mathbf{z}$};
% Text Node
\draw (417,165) node [anchor=north west][inner sep=0.75pt]  [font=\huge]  {$\tilde{\mathbf{z}}$};
% Text Node
\draw (235,350) node [anchor=north west][inner sep=0.75pt]  [font=\huge,rotate=-270] [align=left] {$\displaystyle f(x)$};
% Text Node
\draw (360,398) node [anchor=north west][inner sep=0.75pt]  [rotate=-270] [align=left] {\begin{minipage}[lt]{101.77pt}\setlength\topsep{0pt}
\begin{center}
{\LARGE Downstream}\\{\LARGE Task}
\end{center}

\end{minipage}};
% Text Node
\draw (300,420) node [anchor=north west][inner sep=0.75pt]  [font=\huge] [align=left] {Inferencing};
% Text Node
\draw (478,277) node [anchor=north west][inner sep=0.75pt]   [align=left] {{\Large 0.99}};
% Text Node
\draw (524,328) node [anchor=north west][inner sep=0.75pt]   [align=left] {{\Large 0.52}};
% Text Node
\draw (440,246) node [anchor=north west][inner sep=0.75pt]   [align=left] {{\huge Score}};

\end{tikzpicture}

%% file: images/generation_causal.tikz
\tikzset{every picture/.style={line width=0.75pt}} %set default line width to 0.75pt        

\begin{tikzpicture}[x=0.75pt,y=0.75pt,yscale=-1,xscale=1]
%uncomment if require: \path (0,406); %set diagram left start at 0, and has height of 406

%Shape: Circle [id:dp7457880050294141] 
\draw   (256,112) .. controls (256,98.19) and (267.19,87) .. (281,87) .. controls (294.81,87) and (306,98.19) .. (306,112) .. controls (306,125.81) and (294.81,137) .. (281,137) .. controls (267.19,137) and (256,125.81) .. (256,112) -- cycle ;
%Shape: Circle [id:dp23794232143326854] 
\draw   (316,111) .. controls (316,97.19) and (327.19,86) .. (341,86) .. controls (354.81,86) and (366,97.19) .. (366,111) .. controls (366,124.81) and (354.81,136) .. (341,136) .. controls (327.19,136) and (316,124.81) .. (316,111) -- cycle ;
%Shape: Circle [id:dp6529584559666668] 
\draw   (374,110) .. controls (374,96.19) and (385.19,85) .. (399,85) .. controls (412.81,85) and (424,96.19) .. (424,110) .. controls (424,123.81) and (412.81,135) .. (399,135) .. controls (385.19,135) and (374,123.81) .. (374,110) -- cycle ;
%Shape: Circle [id:dp465073039215554] 
\draw   (316,177) .. controls (316,163.19) and (327.19,152) .. (341,152) .. controls (354.81,152) and (366,163.19) .. (366,177) .. controls (366,190.81) and (354.81,202) .. (341,202) .. controls (327.19,202) and (316,190.81) .. (316,177) -- cycle ;
%Straight Lines [id:da7502121108157895] 
\draw    (281,137) -- (314.68,175.49) ;
\draw [shift={(316,177)}, rotate = 228.81] [color={rgb, 255:red, 0; green, 0; blue, 0 }  ][line width=0.75]    (10.93,-3.29) .. controls (6.95,-1.4) and (3.31,-0.3) .. (0,0) .. controls (3.31,0.3) and (6.95,1.4) .. (10.93,3.29)   ;
%Straight Lines [id:da4670253986820383] 
\draw    (341,136) -- (341,150) ;
\draw [shift={(341,152)}, rotate = 270] [color={rgb, 255:red, 0; green, 0; blue, 0 }  ][line width=0.75]    (10.93,-3.29) .. controls (6.95,-1.4) and (3.31,-0.3) .. (0,0) .. controls (3.31,0.3) and (6.95,1.4) .. (10.93,3.29)   ;
%Straight Lines [id:da07308781783020524] 
\draw    (399,135) -- (367.24,175.43) ;
\draw [shift={(366,177)}, rotate = 308.16] [color={rgb, 255:red, 0; green, 0; blue, 0 }  ][line width=0.75]    (10.93,-3.29) .. controls (6.95,-1.4) and (3.31,-0.3) .. (0,0) .. controls (3.31,0.3) and (6.95,1.4) .. (10.93,3.29)   ;
%Shape: Circle [id:dp1301647869185487] 
\draw   (316,44) .. controls (316,30.19) and (327.19,19) .. (341,19) .. controls (354.81,19) and (366,30.19) .. (366,44) .. controls (366,57.81) and (354.81,69) .. (341,69) .. controls (327.19,69) and (316,57.81) .. (316,44) -- cycle ;
%Straight Lines [id:da8790111310773516] 
\draw    (341,69) -- (341,84) ;
\draw [shift={(341,86)}, rotate = 270] [color={rgb, 255:red, 0; green, 0; blue, 0 }  ][line width=0.75]    (10.93,-3.29) .. controls (6.95,-1.4) and (3.31,-0.3) .. (0,0) .. controls (3.31,0.3) and (6.95,1.4) .. (10.93,3.29)   ;
%Straight Lines [id:da29970367408879994] 
\draw    (341,69) -- (282.92,86.43) ;
\draw [shift={(281,87)}, rotate = 343.3] [color={rgb, 255:red, 0; green, 0; blue, 0 }  ][line width=0.75]    (10.93,-3.29) .. controls (6.95,-1.4) and (3.31,-0.3) .. (0,0) .. controls (3.31,0.3) and (6.95,1.4) .. (10.93,3.29)   ;
%Shape: Circle [id:dp20643482304954186] 
\draw   (316,240) .. controls (316,226.19) and (327.19,215) .. (341,215) .. controls (354.81,215) and (366,226.19) .. (366,240) .. controls (366,253.81) and (354.81,265) .. (341,265) .. controls (327.19,265) and (316,253.81) .. (316,240) -- cycle ;
%Straight Lines [id:da6637677848707277] 
\draw    (341,202) -- (341,213) ;
\draw [shift={(341,215)}, rotate = 270] [color={rgb, 255:red, 0; green, 0; blue, 0 }  ][line width=0.75]    (10.93,-3.29) .. controls (6.95,-1.4) and (3.31,-0.3) .. (0,0) .. controls (3.31,0.3) and (6.95,1.4) .. (10.93,3.29)   ;

% Text Node
\draw (280,112) node   [align=left] {\begin{minipage}[lt]{43.52pt}\setlength\topsep{0pt}
\begin{center}
{\Large $\displaystyle V_{1}$}
\end{center}

\end{minipage}};
% Text Node
\draw (340,112) node   [align=left] {\begin{minipage}[lt]{31.28pt}\setlength\topsep{0pt}
\begin{center}
{\Large $\displaystyle V_{2}$}
\end{center}

\end{minipage}};
% Text Node
\draw (400,112) node   [align=left] {\begin{minipage}[lt]{26.52pt}\setlength\topsep{0pt}
\begin{center}
{\huge $\displaystyle c$}
\end{center}

\end{minipage}};
% Text Node
\draw (340,175) node   [align=left] {\begin{minipage}[lt]{21.76pt}\setlength\topsep{0pt}
\begin{center}
{\huge $\displaystyle z$}
\end{center}

\end{minipage}};
% Text Node
\draw (340,45) node   [align=left] {\begin{minipage}[lt]{31.96pt}\setlength\topsep{0pt}
\begin{center}
{\Large $\displaystyle V_{3}$}
\end{center}

\end{minipage}};
% Text Node
\draw (340,240) node   [align=left] {\begin{minipage}[lt]{21.76pt}\setlength\topsep{0pt}
\begin{center}
{\huge $\displaystyle x$}
\end{center}

\end{minipage}};

\end{tikzpicture}

%% file: images/causal3dident.tikz
\tikzset{every picture/.style={line width=0.75pt}} %set default line width to 0.75pt        

\begin{tikzpicture}[x=0.75pt,y=0.75pt,yscale=-1,xscale=1]
%uncomment if require: \path (0,300); %set diagram left start at 0, and has height of 300

%Shape: Circle [id:dp5811487465385257] 
\draw   (80,67) .. controls (80,46.57) and (96.57,30) .. (117,30) .. controls (137.43,30) and (154,46.57) .. (154,67) .. controls (154,87.43) and (137.43,104) .. (117,104) .. controls (96.57,104) and (80,87.43) .. (80,67) -- cycle ;
%Shape: Circle [id:dp26297852421621526] 
\draw   (170,67) .. controls (170,46.57) and (186.57,30) .. (207,30) .. controls (227.43,30) and (244,46.57) .. (244,67) .. controls (244,87.43) and (227.43,104) .. (207,104) .. controls (186.57,104) and (170,87.43) .. (170,67) -- cycle ;
%Shape: Circle [id:dp35410989451625374] 
\draw   (260,67) .. controls (260,46.57) and (276.57,30) .. (297,30) .. controls (317.43,30) and (334,46.57) .. (334,67) .. controls (334,87.43) and (317.43,104) .. (297,104) .. controls (276.57,104) and (260,87.43) .. (260,67) -- cycle ;
%Shape: Circle [id:dp8626574612773585] 
\draw   (350,67) .. controls (350,46.57) and (366.57,30) .. (387,30) .. controls (407.43,30) and (424,46.57) .. (424,67) .. controls (424,87.43) and (407.43,104) .. (387,104) .. controls (366.57,104) and (350,87.43) .. (350,67) -- cycle ;
%Shape: Circle [id:dp790433915900072] 
\draw   (80,157) .. controls (80,136.57) and (96.57,120) .. (117,120) .. controls (137.43,120) and (154,136.57) .. (154,157) .. controls (154,177.43) and (137.43,194) .. (117,194) .. controls (96.57,194) and (80,177.43) .. (80,157) -- cycle ;
%Shape: Circle [id:dp04536234641762782] 
\draw   (170,157) .. controls (170,136.57) and (186.57,120) .. (207,120) .. controls (227.43,120) and (244,136.57) .. (244,157) .. controls (244,177.43) and (227.43,194) .. (207,194) .. controls (186.57,194) and (170,177.43) .. (170,157) -- cycle ;
%Shape: Circle [id:dp17993210993400277] 
\draw   (260,157) .. controls (260,136.57) and (276.57,120) .. (297,120) .. controls (317.43,120) and (334,136.57) .. (334,157) .. controls (334,177.43) and (317.43,194) .. (297,194) .. controls (276.57,194) and (260,177.43) .. (260,157) -- cycle ;
%Straight Lines [id:da4638596134033006] 
\draw    (117,104) -- (117,117) ;
\draw [shift={(117,120)}, rotate = 270] [fill={rgb, 255:red, 0; green, 0; blue, 0 }  ][line width=0.08]  [draw opacity=0] (8.93,-4.29) -- (0,0) -- (8.93,4.29) -- cycle    ;
%Straight Lines [id:da25699403279353406] 
\draw    (207,104) -- (207,117) ;
\draw [shift={(207,120)}, rotate = 270] [fill={rgb, 255:red, 0; green, 0; blue, 0 }  ][line width=0.08]  [draw opacity=0] (8.93,-4.29) -- (0,0) -- (8.93,4.29) -- cycle    ;
%Straight Lines [id:da05479886294719427] 
\draw    (297,104) -- (297,117) ;
\draw [shift={(297,120)}, rotate = 270] [fill={rgb, 255:red, 0; green, 0; blue, 0 }  ][line width=0.08]  [draw opacity=0] (8.93,-4.29) -- (0,0) -- (8.93,4.29) -- cycle    ;
%Straight Lines [id:da14452941358609572] 
\draw    (360,93.75) -- (328.97,129.48) ;
\draw [shift={(327,131.75)}, rotate = 310.97] [fill={rgb, 255:red, 0; green, 0; blue, 0 }  ][line width=0.08]  [draw opacity=0] (8.93,-4.29) -- (0,0) -- (8.93,4.29) -- cycle    ;
%Straight Lines [id:da24993107378943424] 
\draw    (178.5,93.25) -- (147.47,128.98) ;
\draw [shift={(145.5,131.25)}, rotate = 310.97] [fill={rgb, 255:red, 0; green, 0; blue, 0 }  ][line width=0.08]  [draw opacity=0] (8.93,-4.29) -- (0,0) -- (8.93,4.29) -- cycle    ;
%Straight Lines [id:da9955139476083812] 
\draw    (236,89.25) -- (268.61,129.42) ;
\draw [shift={(270.5,131.75)}, rotate = 230.93] [fill={rgb, 255:red, 0; green, 0; blue, 0 }  ][line width=0.08]  [draw opacity=0] (8.93,-4.29) -- (0,0) -- (8.93,4.29) -- cycle    ;

% Text Node
\draw (82,50) node [anchor=north west][inner sep=0.75pt]  [font=\huge] [align=left] {$\displaystyle pos_{spl}$};
% Text Node
\draw (175,50) node [anchor=north west][inner sep=0.75pt]  [font=\huge] [align=left] {$\displaystyle class$};
% Text Node
\draw (265,50) node [anchor=north west][inner sep=0.75pt]  [font=\huge] [align=left] {$\displaystyle hue_{bg}$};
% Text Node
\draw (350,50) node [anchor=north west][inner sep=0.75pt]  [font=\huge] [align=left] {$\displaystyle hue_{spl}$};
% Text Node
\draw (82,145) node [anchor=north west][inner sep=0.75pt]  [font=\huge] [align=left] {$\displaystyle pos_{obj}$};
% Text Node
\draw (175,145) node [anchor=north west][inner sep=0.75pt]  [font=\huge] [align=left] {$\displaystyle rot_{obj}$};
% Text Node
\draw (261,145) node [anchor=north west][inner sep=0.75pt]  [font=\huge] [align=left] {$\displaystyle hue_{obj}$};

\end{tikzpicture}

%% file: images/datasets.tikz
\tikzset{every picture/.style={line width=0.75pt}} %set default line width to 0.75pt        

\begin{tikzpicture}[x=0.75pt,y=0.75pt,yscale=-1,xscale=1]
%uncomment if require: \path (0,443); %set diagram left start at 0, and has height of 443

%Image [id:dp1733565750881536] 
\draw (350,322.5) node  {\includegraphics[width=225pt,height=93.75pt]{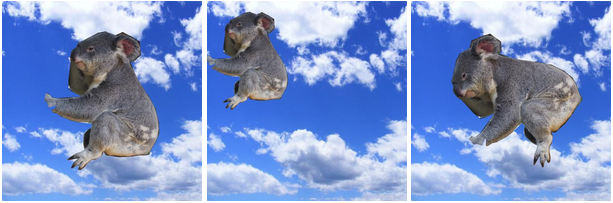}};
%Image [id:dp36546646294668084] 
\draw (350,192.5) node  {\includegraphics[width=225pt,height=93.75pt]{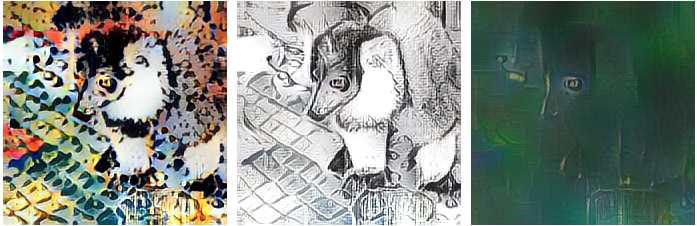}};
%Image [id:dp7747690848294577] 
\draw (350,62.5) node  {\includegraphics[width=225pt,height=93.75pt]{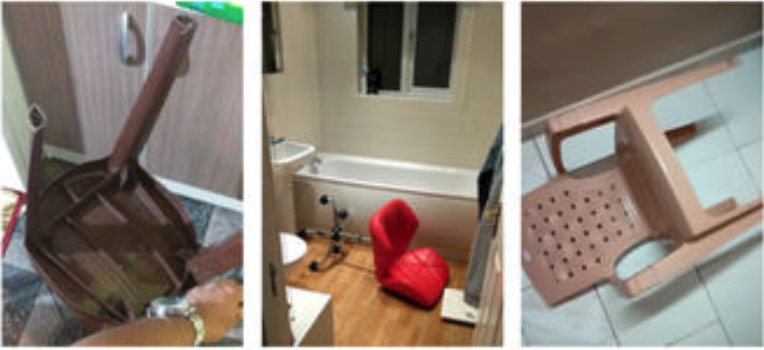}};

\end{tikzpicture}

%% file: appendix.tex
\section{Recent Descriminative SSL Formulations} \label{appendix:ssl formulations}
\begin{table}[htbp]
\resizebox{0.45\textwidth}{!}{
\begin{tabular}{c|c}
\hline
Notation & Description             \\ \hline
   $\mathbf{x}_i$      & An image sample         \\ \hline
 $\mathcal{A}_1(\mathbf{x}_*),\mathcal{A}_2(\mathbf{x}_*)$    & Two random augmentations   \\ \hline
$f_{\theta}(.)$     & A feature encoder parameterized by $\theta$ \\ \hline
$f_{\xi}(.)$  & An EMA encoder parameterized by $\xi$                        \\ \hline
  $\tau$         &     A temperature scaling term        \\ \hline
  $p_{\theta}(.)$ & An MLP predictor parameterized by $\theta$. \\ \hline
  $\texttt{sg}$ & \texttt{stop-gradient} operation. \\ \hline
\end{tabular}}
\caption{Notations for important elements in SSL. \textbf{Note that the extracted or projected representations are normalized to a unit sphere unless specified otherwise.}}
\end{table}
\textbf{SimCLR}\cite{chen2020simple} optimizes InfoNCE to maximize similarity between positive pairs and minimize similarity between negative pairs. Positive pairs are two augmented views of an image sample, $\mathbf{\tilde{x}}_{2i},\mathbf{\tilde{x}}_{2i+1}=\mathcal{A}_1(\mathbf{x}_i),\mathcal{A}_2(\mathbf{x}_i)$. Negative pairs are all other augmented samples in a mini training batch. For a batch with $N$ image samples, the augmentation produces $2N$ augmented samples. A feature encoder $f_{\theta}(.)$ extracts representations of the batch data $\mathbf{\tilde{z}}_i=f_{\theta}(\mathbf{\tilde{x}}_i)$. SimCLR optimizes the following objective:
\begin{equation*}
    \mathcal{L}_{SimCLR}=\mathop{\mathbb{E}}_{i}[-log\frac{e^{\mathbf{\tilde{z}}_{2i}^{\top}\mathbf{\tilde{z}}_{2i+1}/\tau}}{\sum_{j=1,j\neq{2i}}^{2B}{e^{\mathbf{\tilde{z}}_{2i}^{\top}\mathbf{\tilde{z}}_{j}/\tau}}}]
\end{equation*}

\textbf{MoCo}/\textbf{MoCo V2} optimzes InfoNCE to maximize similarity between positive pairs and minimize similarity between negative pairs. Positive pairs are two augmented views of an image sample, $\mathbf{\tilde{x}}_{2i},\mathbf{\tilde{x}}_{2i+1}=\mathcal{A}_1(\mathbf{x}_i),\mathcal{A}_2(\mathbf{x}_i)$. But negative pairs are representations learned via a moving-averaged network, $f_{\xi}(.)$, and stored in a memory bank with size $K$, $\mathcal{M}_K$. And $\xi=m\xi+(1-m)\theta$, where $m$ is a momentum coefficient. $\mathbf{\tilde{z}}_{2i}=f_{\theta}(\mathbf{\tilde{x}}_{2i})$, $\mathbf{\hat{z}}_{2i+1}=f_{\xi}(\mathbf{\tilde{x}}_{2i+1})$, $\mathbf{\hat{z}}_j=f_{\xi}(\mathbf{\tilde{x}}_j) \in \mathcal{M}_K$. MoCo optimizes the following objective:
\begin{equation*}
    \mathcal{L}_{MoCo}=\mathop{\mathbb{E}}_{i}[-log\frac{e^{\mathbf{\tilde{z}}_{2i}^{\top}\mathbf{\hat{z}}_{2i+1}/\tau}}{\sum_{j=1}^{K}{e^{\mathbf{\tilde{z}}_{2i}^{\top}\mathbf{\hat{z}}_{j}/\tau}}}]
\end{equation*}

\textbf{BYOL} aligns the projection of a representation of an augmented sample with an EMA representation of another augmented sample. The main difference between BYOL and SimCLR/MoCo is the claim that BYOL only formulates the objective on positive pairs. $\mathbf{\tilde{x}}_{2i},\mathbf{\tilde{x}}_{2i+1}=\mathcal{A}_1(\mathbf{x}_i),\mathcal{A}_2(\mathbf{x}_i)$. An MLP predictor, $p_{\theta}(.)$, further projects the representation extracted by the feature encoder to an embedding space and the \textit{EMA} representation predicts the projected embedding/representation by alignment. $\mathbf{\tilde{z}}_{2i}=f_{\theta}(\mathbf{\tilde{x}}_{2i})$,$\mathbf{\hat{z}}_{2i+1}=f_{\xi}(\mathbf{\tilde{x}}_{2i+1})$. BYOL optimizes the following objective:

\begin{equation*}
    \mathcal{L}_{BYOL}=\mathop{\mathbb{E}}_{i}\|p_{\theta}(\mathbf{\tilde{z}}_{2i})-\mathbf{\hat{z}}_{2i+1}\|^2_2
\end{equation*}

\textbf{SimSiam} aligns the projection of a representation of an augmented sample with a detached representation of another augmented sample. Unlike BYOL or MoCo, SiamSiam omits the EMA encoder that the author deem to be unnecessary for a stable representation learning. An MLP predictor, $p_{\theta}(.)$, further projects the representation extracted by the feature encoder to an embedding space and the \textit{detached} representation predicts the projected embedding/representation by alignment. $\mathbf{\tilde{z}}_{2i}=f_{\theta}(\mathbf{\tilde{x}}_{2i})$,$\mathbf{\hat{z}}_{2i+1}=f_{\theta}(\mathbf{\tilde{x}}_{2i+1})$\footnote{The $\mathbf{z}$ is not normalized yet, since in the loss function both projected and extracted representations are normalized.}. SimSiam optimizes the following objective:

\begin{equation*}
    \mathcal{L}_{SimSiam}=\mathop{\mathbb{E}}_{i}[-\frac{p_{\theta}(\mathbf{\tilde{z}}_{2i})}{\|p_{\theta}(\mathbf{\tilde{z}}_{2i})\|_2}\cdot{\frac{\mathbf{\hat{z}}_{2i+1}}{\|\mathbf{\hat{z}}_{2i+1}\|_2}}]
\end{equation*}

\textbf{Barlow Twins} aligns positive representations of two augmented samples in feature dimensions and reduces redundancy cross different feature dimensions. Different to all aforementioned SSL methods, the authors suggest to standard normalize the representation (zero mean and unit std) instead of unit sphere normalization. However, as stated in the paper, either normalization scheme works under Barlow Twin method. $\mathcal{Z}^A=\{\mathbf{\tilde{z}}_{2i}\}_{i=1}^N=\{f_{\theta}(\mathbf{\tilde{x}}_{2i})\}_{i=1}^N$,$\mathcal{Z}^B=\{\mathbf{\tilde{z}}_{2i+1}\}_{i=1}^N=\{f_{\theta}(\mathbf{\tilde{x}}_{2i+1})\}_{i=1}^N$. And $\mathcal{Z}^A$ and $\mathcal{Z}^B$ are normalized over the batch statistics. Barlow Twins optimizes the following objective:
\begin{equation} \label{eq:barlow}
    \mathcal{L}_{BarlowTwins} = \mathop{\sum}_{a}(1-\mathcal{C}_{aa})^2+\lambda\mathop{\sum}_{a}\mathop{\sum}_{b\neq{a}}{\mathcal{C}_{ab}}^2
\end{equation}
\begin{equation*}
    \mathcal{C}_{ab} = \frac{\sum_{i=1}^N[\mathbf{\tilde{z}}_{2i}]_a [\mathbf{\tilde{z}}_{2i+1}]_b}{\sqrt{\sum_i([\mathbf{\tilde{z}}_{2i}]_a)^2}\sqrt{\sum_i([\mathbf{\tilde{z}}_{2i+1}]_b)^2}}
\end{equation*}
\section{Extended Theory and Proofs} \label{appendix:extend}
In sections \ref{formulation} and \ref{alignment and regularization} we introduce our data generation process and prove that all SSL methods benefit from the alignment term in their objectives. Here we extend the theory to include the output entropy(s) of the encoder network(s) and provide analysis on how SSL prevent representation collapse by maximizing the output entropy of the network. 

\begin{theorem}
With a data generation process described in \ref{formulation}, all discriminative SSL objectives have an alignment loss function between positive pairs from the network and output entropy loss function(s) of the network(s):
\begin{equation}
    \mathcal{L}_{SSL} = \|f(\mathbf{x})-f(\mathbf{\tilde{x}})\|^2_2-H(f(\mathbf{x},\theta))
\end{equation} \label{eq:ssl objectives}
\end{theorem}
%The third term in (\ref{eq:ssl objectives}) applies to SSL methods that have an EMA encoder. 

For InfoNCE-driven SSL methods, the proof is:
\begin{align}
    \begin{split}
  \lim_{K\to\infty} \mathcal{L}_{InfoNCE}&-logK = \\
    \underbrace{-\frac{1}{\tau}\mathop{\mathbb{E}}_{(\mathbf{x},\tilde{\mathbf{x}})}[f(\mathbf{x})^{\top}f(\tilde{\mathbf{x}})]}_{alignment}
    &+ \underbrace{\mathop{\mathbb{E}}_{\mathbf{x}}[log\mathop{\mathbb{E}}_{\mathbf{x}^-}[e^{f(\mathbf{x}^-)^{\top}f(\mathbf{x})/\tau}]]}_{uniformity}
\end{split} \label{eq:align and uniform}
\end{align}

with the \textit{alignment} term in (\ref{eq:align and uniform}) equivlent to $\mathop{\mathbb{E}}_{(\mathbf{x},\tilde{\mathbf{x}})}(1-\|f(\mathbf{x})-f(\tilde{\mathbf{x}})\|^2_2/2)$ and the \textit{uniformity} term equivalent to $-H(f(\mathbf{x}))+log{C_q(\mathbf{z})}$ \cite{ahmad1976nonparametric}. Hence complete the proof by:

\begin{align}
    \begin{split}
        \lim_{K\to\infty} \mathcal{L}_{InfoNCE}&-logK = \\
        \frac{1}{\tau}\mathop{\mathbb{E}}_{(\mathbf{x},\tilde{\mathbf{x}})}[\|f(\mathbf{x})-f(\tilde{\mathbf{x}})\|^2_2/2-1] &- H(f(\mathbf{x}))+log{C_q(\mathbf{z})}
    \end{split} \label{eq:infonce reform}
\end{align}

% \begin{align*}
%     \begin{split}
%         \lim_{K\to\infty} \mathcal{L}_{InfoNCE}-logK &-log{C_q(\mathbf{z})} = \\
%         \frac{1}{\tau}\mathop{\mathbb{E}}_{(\mathbf{x},\tilde{\mathbf{x}})}[\|f(\mathbf{x})-f(\tilde{\mathbf{x}})\|^2_2/2-1] &- H(f(\mathbf{x}))
%     \end{split} \label{eq:infonce reform}
% \end{align*}

For both EMA-driven and Siamese with predictor SSLs, we show that the loss function can be reformulated to three terms that first two are the alignment between positive pairs through the online/trainable network, and the alignment between same data sample from two networks. The third term can be further approximated via second order Taylor expansion around $\mathbf{z}$.

\begin{equation}
    \begin{split}
    - 2\mathop{\mathbb{E}}_{(\mathbf{x},\tilde{\mathbf{x}})}[(p'(\tilde{\mathbf{x}},\theta)-p'(\mathbf{x},\theta))^{\top}(p'(\tilde{\mathbf{x}},\theta)-f(\tilde{\mathbf{x}},\xi))] \\
     = -2(1-\mathop{\mathbb{E}}_{(\mathbf{x},\tilde{\mathbf{x}})}[p'(\mathbf{x},\theta)^{\top}p'(\tilde{\mathbf{x}},\theta)]-\\
    \mathop{\mathbb{E}}_{\mathbf{\tilde{x}}}[p'(\tilde{\mathbf{x}},\theta)^{\top}f(\tilde{\mathbf{x}},\xi))]+\mathop{\mathbb{E}}_{(\mathbf{x},\tilde{\mathbf{x}})}[p'(\mathbf{x},\theta)^{\top}f(\tilde{\mathbf{x}},\xi))]) \\
    \approx -2(1-log(\mathop{\mathbb{E}}_{(\mathbf{x},\tilde{\mathbf{x}})}[e^{p'(\mathbf{x},\theta)^{\top}p'(\tilde{\mathbf{x}},\theta)}])+\frac{\mathbb{V}[p'(\mathbf{x},\theta)^{\top}p'(\tilde{\mathbf{x}},\theta)]}{2\mathbb{E}[p'(\mathbf{x},\theta)^{\top}p'(\tilde{\mathbf{x}},\theta)]^2} \\
    -log(\mathop{\mathbb{E}}_{\tilde{\mathbf{x}}}[e^{p'(\tilde{\mathbf{x}},\theta)^{\top}f(\tilde{\mathbf{x}},\xi))}])+\frac{\mathbb{V}[p'(\tilde{\mathbf{x}},\theta)^{\top}f(\tilde{\mathbf{x}},\xi))]}{2\mathbb{E}[p'(\tilde{\mathbf{x}},\theta)^{\top}f(\tilde{\mathbf{x}},\xi))]^2} \\
    +log(\mathop{\mathbb{E}}_{(\mathbf{x},\tilde{\mathbf{x}})}[e^{p'(\mathbf{x},\theta)^{\top}f(\tilde{\mathbf{x}},\xi))}])-\frac{\mathbb{V}[p'(\mathbf{x},\theta)^{\top}f(\tilde{\mathbf{x}},\xi))]}{2\mathbb{E}[p'(\mathbf{x},\theta)^{\top}f(\tilde{\mathbf{x}},\xi))]^2})
    \end{split}
\end{equation}

Assuming that $\kappa_2$ is a large number (as set by $1/\tau$ in SimCLR and MoCo), then the variance terms $\mathbb{V}[.]\approx{0}$. 

\begin{equation} \label{eq:ema reform}
    \begin{split}
    \mathcal{L} = \mathop{\mathbb{E}}_{(\mathbf{x},\tilde{\mathbf{x}})}\|p'(\mathbf{x},\theta)-p'(\tilde{\mathbf{x}},\theta)+p'(\tilde{\mathbf{x}},\theta)-f(\tilde{\mathbf{x}},\xi)\|^2_2 \\
    = \mathop{\mathbb{E}}_{(\mathbf{x},\tilde{\mathbf{x}})}\|p'(\mathbf{x},\theta)-p'(\tilde{\mathbf{x}},\theta)\|^2_2+\mathop{\mathbb{E}}_{\tilde{\mathbf{x}}}\|p'(\tilde{\mathbf{x}},\theta)-f(\tilde{\mathbf{x}},\xi)\|^2_2 \\
    - 2\mathop{\mathbb{E}}_{(\mathbf{x},\tilde{\mathbf{x}})}[(p'(\tilde{\mathbf{x}},\theta)-p'(\mathbf{x},\theta))^{\top}(p'(\tilde{\mathbf{x}},\theta)-f(\tilde{\mathbf{x}},\xi))] \\
    \approx \mathop{\mathbb{E}}_{(\mathbf{x},\tilde{\mathbf{x}})}\|p'(\mathbf{x},\theta)-p'(\tilde{\mathbf{x}},\theta)\|^2_2+\mathop{\mathbb{E}}_{\tilde{\mathbf{x}}}\|p'(\tilde{\mathbf{x}},\theta)-f(\tilde{\mathbf{x}},\xi)\|^2_2 \\
    -2+2log(\mathop{\mathbb{E}}_{(\mathbf{x},\tilde{\mathbf{x}})}[e^{p'(\mathbf{x},\theta)^{\top}p'(\tilde{\mathbf{x}},\theta)}])+2log(\mathop{\mathbb{E}}_{\tilde{\mathbf{x}}}[e^{p'(\tilde{\mathbf{x}},\theta)^{\top}f(\tilde{\mathbf{x}},\xi))}])\\
    -2log(\mathop{\mathbb{E}}_{(\mathbf{x},\tilde{\mathbf{x}})}[e^{p'(\mathbf{x},\theta)^{\top}f(\tilde{\mathbf{x}},\xi))}]) \\
    = \mathop{\mathbb{E}}_{(\mathbf{x},\tilde{\mathbf{x}})}\|p'(\mathbf{x},\theta)-p'(\tilde{\mathbf{x}},\theta)\|^2_2+\mathop{\mathbb{E}}_{\tilde{\mathbf{x}}}\|p'(\tilde{\mathbf{x}},\theta)-f(\tilde{\mathbf{x}},\xi)\|^2_2 \\
    -2-2H(p'(\mathbf{x}))+2logC_q(\mathbf{z})+2log(\mathop{\mathbb{E}}_{\tilde{\mathbf{x}}}[e^{p'(\tilde{\mathbf{x}},\theta)^{\top}f(\tilde{\mathbf{x}},\xi))}])\\
    -2log(\mathop{\mathbb{E}}_{(\mathbf{x},\tilde{\mathbf{x}})}[e^{p'(\mathbf{x},\theta)^{\top}f(\tilde{\mathbf{x}},\xi))}])
    \end{split}
\end{equation} 

Since $f(\mathbf{x},\theta)$ and $f(\tilde{\mathbf{x}},\xi)$ maps in the same space $\mathbb{R}^{d_2}$, $p$ can be considered as a bijective linear transformation within $\mathbb{R}^{b_2}$. In \cite{DBLP:journals/corr/abs-1109-4856} by change of variable: $H(\mathbf{Y})=H(\mathbf{X})+\mathbb{E}[log|\mathbf{J}_m|]-H(\mathbf{X}|\mathbf{Y})$ if $\mathbf{Y}=m\mathbf{X}$, where $m$ is a projection matrix mapping $\mathbf{X}\to \mathbf{Y}$ and $\mathbf{J}_m$ is the Jacobian of $m$, $\frac{\delta{m}}{\delta{\mathbf{x}}}$.
This relates the last two terms in (\ref{eq:ema reform}) to maximizing the output cross entropy of $p'$ and $f$ w.r.t the same sample, and minimizing the output cross entropy of $p'$ and $f$ w.r.t positive samples. This also hints on the importance of the predictor in BYOL, since removing the $p$ in $p\circ f$, $2log(\mathop{\mathbb{E}}_{(\mathbf{x},\tilde{\mathbf{x}})}[e^{p'(\mathbf{x},\theta)^{\top}p'(\tilde{\mathbf{x}},\theta)}])$ and $-2log(\mathop{\mathbb{E}}_{(\mathbf{x},\tilde{\mathbf{x}})}[e^{p'(\mathbf{x},\theta)^{\top}f(\tilde{\mathbf{x}},\xi))}])$ will cancel out and omit the output entropy maximization objective resulting in the representation collapse.

The same analysis applies to Siamese with predictor SSL. In case the predictor and/or \texttt{stop-gradient} is removed, the output entropy maximization objective will be no longer available and lead to a trivial solution.

For Barlow Twins, the objective can be regarded as minimizing the information loss between two augmented examples. In Appendix.A in \cite{zbontar2021barlow}, the author formulate such relation to the Information Bottleneck Principle:
\begin{equation} \label{eq:barlow reform}
    \begin{split}
        \mathcal{IB}_{\theta} &= I(f_{\theta}(\mathbf{x}),\mathbf{x})-\beta I(f_{\theta}(\mathbf{x}),\mathbf{\tilde{x}}) \\
         &= H(f_{\theta}(\mathbf{x})|\mathbf{x})+\frac{1-\beta}{\beta}H(f_{\theta}(\mathbf{x}))
    \end{split}
\end{equation}

The first term in (\ref{eq:barlow reform}) is linked to the alignment term in (\ref{eq:barlow}) when \cite{zbontar2021barlow} assumes that $f(\mathbf{x})$ follows a Gaussian distribution. However, in our representation learning formulation, we assume the conditional distribution of positive pairs follows a vMF distribution (\ref{eq:representation}). We can further decompose (\ref{eq:barlow reform}):
\begin{equation} \label{eq:barlow reform2}
    \begin{split}
        \mathcal{IB}_{\theta} &= \mathcal{L}_{alignment}-H(f_{\theta}(\mathbf{x}))+\frac{1-\beta}{\beta}H(f_{\theta}(\mathbf{x}))
    \end{split}
\end{equation}
As suggested in \cite{zbontar2021barlow}, when $\beta<1$ the best solution of (\ref{eq:barlow reform2}) is to set the representation to a constant, i.e. representation collapse. When $\beta>1$, the last term in (\ref{eq:barlow reform2}) is the same as maximizing the output entropy of the network.  

Once we prove that all SSL objectives contain the alignment term and output entropy maximization term, we demonstrate that the cross entropy between $p(.|\mathbf{z})$ in (\ref{eq:generation2}) and $q_h(.|\mathbf{z})$ in (\ref{eq:representation}) can be formulated with the $\mathcal{L}_{alignmet}-H(f_{\theta}(\mathbf{x}))$ as illustrated in \cite{Zimmermann2021}.

\begin{equation}
    \begin{split}
        \mathop{\mathbb{E}}_{\mathbf{z}\sim p(\mathbf{z})}[H(p(.|\mathbf{z}),q_h(.|\mathbf{z}))] \\
        = - \mathop{\mathbb{E}}_{\mathbf{z}\sim p(\mathbf{z})}[\mathop{\mathbb{E}}_{\mathbf{\tilde{z}}\sim p(\mathbf{\tilde{z}}|\mathbf{z})}[log(q_h(\mathbf{\tilde{z}}|\mathbf{z}))]] \\
        = \kappa_2 \mathop{\mathbb{E}}_{(\mathbf{z},\mathbf{\tilde{z}})}[h(\mathbf{\tilde{z}})^{\top}h(\mathbf{z})]-\mathop{\mathbb{E}}_{\mathbf{z}}[log(\mathop{\mathbb{E}}_{\mathbf{\tilde{z}}}[e^{h(\mathbf{\tilde{z}})^{\top}h(\mathbf{z})\kappa_1}])] \\
        = \mathcal{L}_{alignment}-H(h(\mathbf{z}))
    \end{split}
\end{equation}
Since $H(.)\leq 0$, then the alignment loss will be a lower bound for $\mathop{\mathbb{E}}_{\mathbf{z}\sim p(\mathbf{z})}[H(p(.|\mathbf{z}),q_h(.|\mathbf{z}))]$.

Finally, we can use \textbf{Proposition 1} and \textbf{Proposition 2} in \cite{Zimmermann2021} to prove \ref{thm:match} and \ref{thm:matrix}.
\section{Causal3DIdent} \label{appendix:causal3d}
\begin{figure*}[htbp]
    \centering
    \includegraphics[scale=0.45]{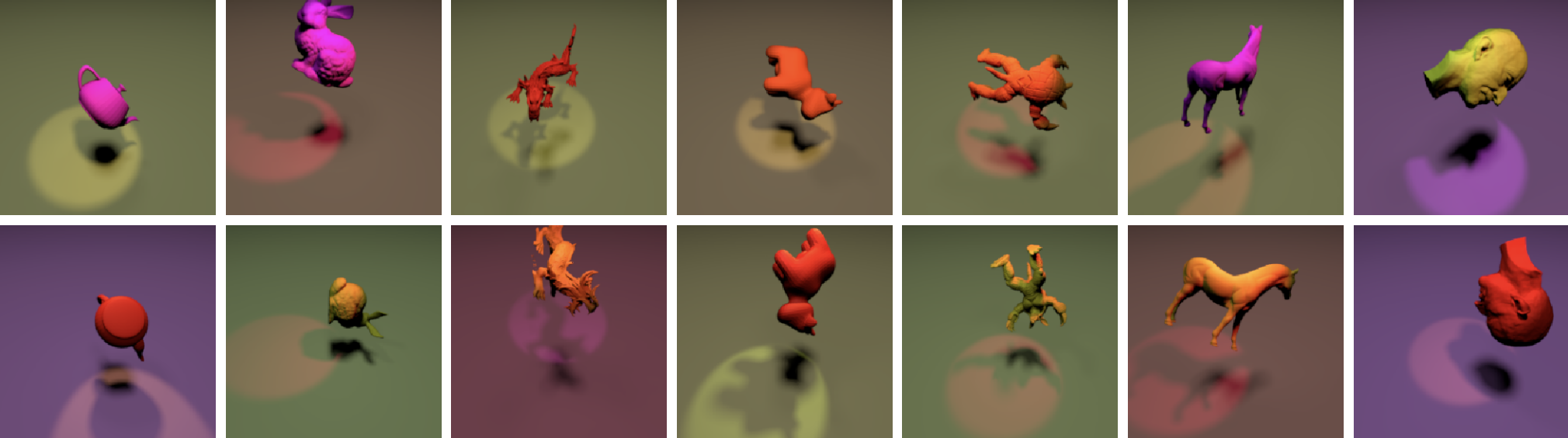}
    \caption{Examples of images in Causal3DIdent. Each image is associated with a 10-dimensional ground-truth latent representation. [$pos_{obj}=(x,y,z)$, $rot_{obj}=(\phi,\theta,\psi)$, $hue_{obj}$, $pos_{spl}$, $hue_{spl}$, $hue_{bg}$]}
    \label{fig:causal3d_dataset}
\end{figure*}
3DIdent contained 7 object classes: Teapot, Hare, Dragon, Cow, Armadillo, Horse, Head. For spotlight position, spotlight hue, and background hue, variable values are sampled from $U(-1,1)$. The dependence is imposed by varying the mean ($\mu$) of a truncated normal distribution with standard deviation $\sigma=0.5$, truncated to the range $[-1,1]$. See Appendix B in \cite{Von2021} for the dependency on $\mu$.

\begin{figure}[htbp]
    \centering
    \begin{subfigure}[b]{0.48\textwidth}
          \centering
          \resizebox{\linewidth}{!}{\includegraphics{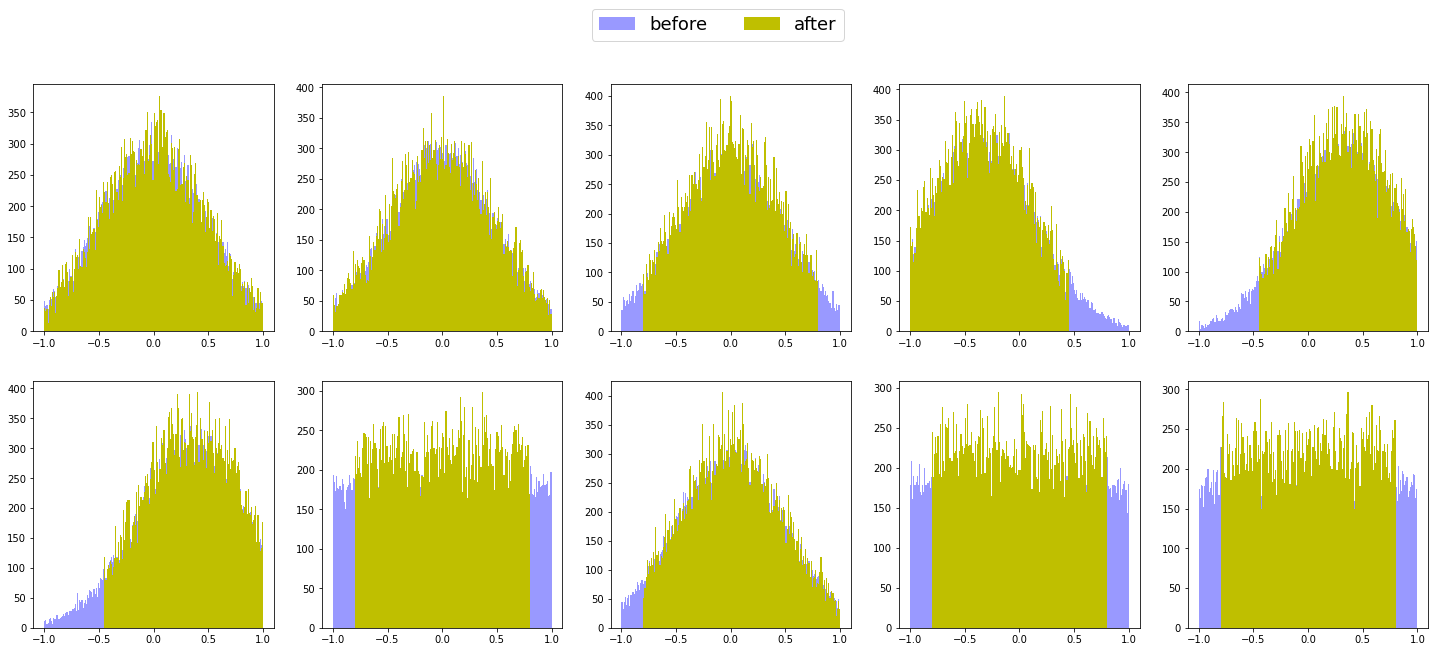}}
          \caption{Teapot}
          \label{fig:teapot}
     \end{subfigure}
     \begin{subfigure}[b]{0.48\textwidth}
          \centering
          \resizebox{\linewidth}{!}{\includegraphics{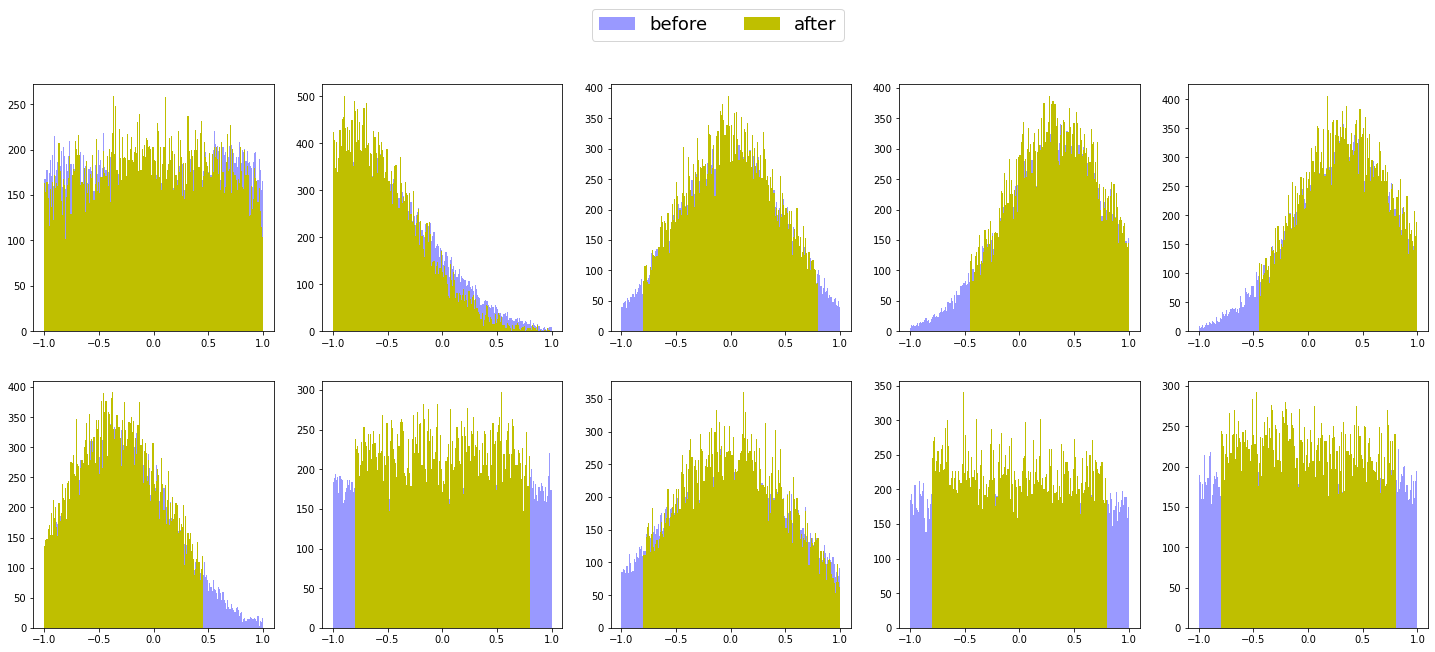}}  
          \caption{Dragon}
          \label{fig:dragon}
     \end{subfigure}
     \begin{subfigure}[b]{0.48\textwidth}
          \centering
          \resizebox{\linewidth}{!}{\includegraphics{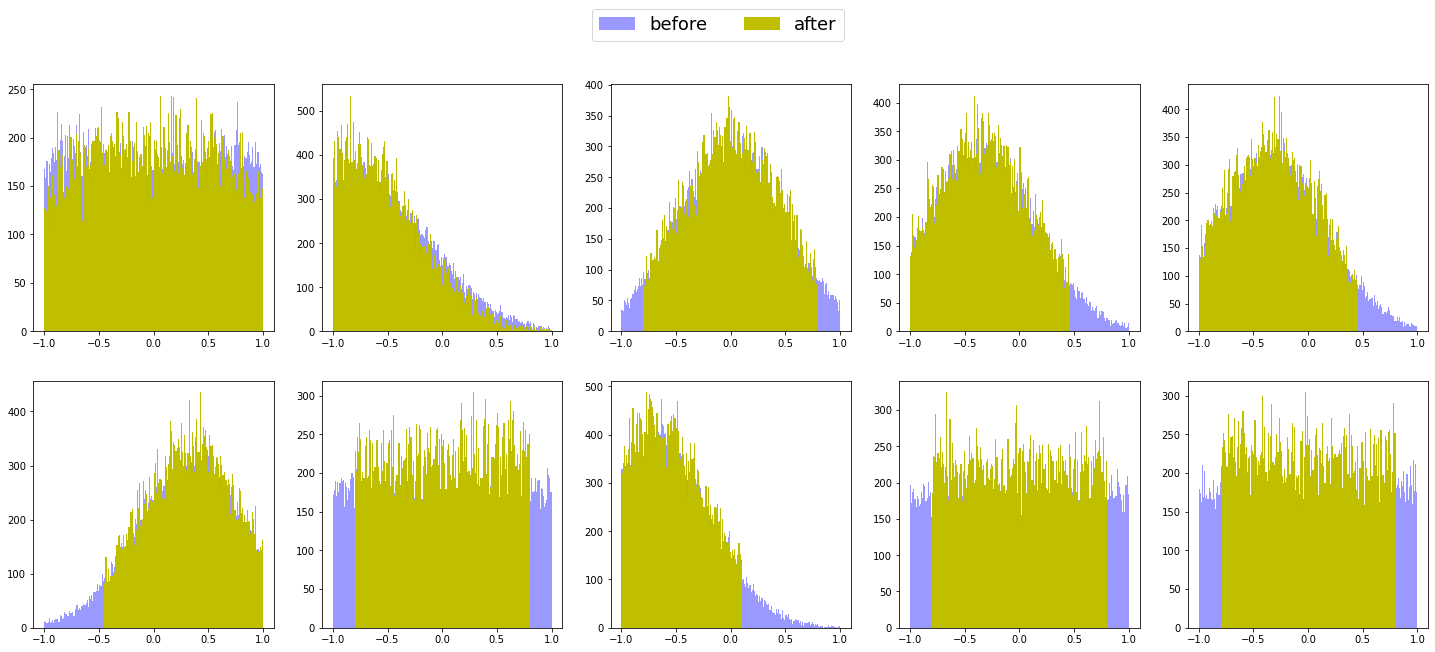}}
          \caption{Horse}
          \label{fig:horse}
     \end{subfigure}
    \caption{Examples of sampling intervened data samples. \textcolor{blue}{Before} sampling and \textcolor{green}{After} sampling.}
    \label{fig:causal3d_sampling}
\end{figure}

To exclude variable range with extreme values, we exclude edges for uniformly sampled variables ($\leq -0.8$ and $\geq 0.8$), and exclude smaller portion tail for dependent variables ($\leq \mu-0.8$ if $\mu>0$,$\geq \mu+0.8$ if $\mu\geq 0$). To ensure the training data size is the same after sampling, the included samples are delicately sampled to match the original data size. Three illustrative examples are shown in Figure \ref{fig:causal3d_sampling}.

This section contains the values of hyperparameters for SSL methods. \textbf{Note the batch size is set to 128}.
\begin{table}[htbp]
\centering
\begin{tabular}{|c|c|}
\hline
                                                       & Hyperparameters \\ \hline
SimCLR                                                 &      $\tau=0.07$           \\ \hline
MoCo                                                   &    $K=65536$,$\tau=0.07$,$\alpha=0.99$          \\ \hline
BYOL                                                   &      $\alpha=0.99$           \\ \hline
SimSiam                                                &      None           \\ \hline
\begin{tabular}[c]{@{}c@{}}Barlow Twins\end{tabular} &      $\lambda=0.005$       \\ \hline
\end{tabular}
\caption{Hyperparameters for SSL methods in training Causal3dIdent.}
\end{table}

To verify that the difference between seen and unseen distribution does not induce large discrepancy in the performance, we evaluate the accuracy on all classes and report the difference between seen and unseen distributions. See Table \ref{tab:difference} for results. On average, the difference is accuracy is about $2\%$ to $3\%$, which is not comparable to the minimum reduction in accuracy $\approx20\%$ reported in Figure \ref{fig:causal_parents}.

\begin{table}[htbp]
    \centering
    \begin{tabular}{lrrrrr}
        \toprule
        {} &  simclr &   moco &   byol &  simsiam &  barlow \\
        \midrule
        acc &  0.0343 & 0.0370 & 0.0190 &   0.0168 &  0.0147 \\
        0   &  0.0860 & 0.0860 & 0.0392 &   0.0279 &  0.0335 \\
        1   &  0.0761 & 0.0866 & 0.0476 &   0.0320 &  0.0334 \\
        2   &  0.0241 & 0.0179 & 0.0178 &   0.0097 &  0.0095 \\
        3   &  0.0445 & 0.0449 & 0.0164 &   0.0261 &  0.0090 \\
        4   &  0.0543 & 0.0458 & 0.0300 &   0.0226 &  0.0126 \\
        5   &  0.0603 & 0.0488 & 0.0262 &   0.0234 &  0.0281 \\
        6   &  0.0542 & 0.0445 & 0.0420 &   0.0655 &  0.0387 \\
        \bottomrule
        \end{tabular}
    \caption{Accuracy discrepancy between seen and unseen distributions.}
    \label{tab:difference}
\end{table}

\begin{figure}[htbp]
    \centering
    \begin{subfigure}[b]{0.5\textwidth}
          \centering
          \resizebox{\linewidth}{!}{\includegraphics{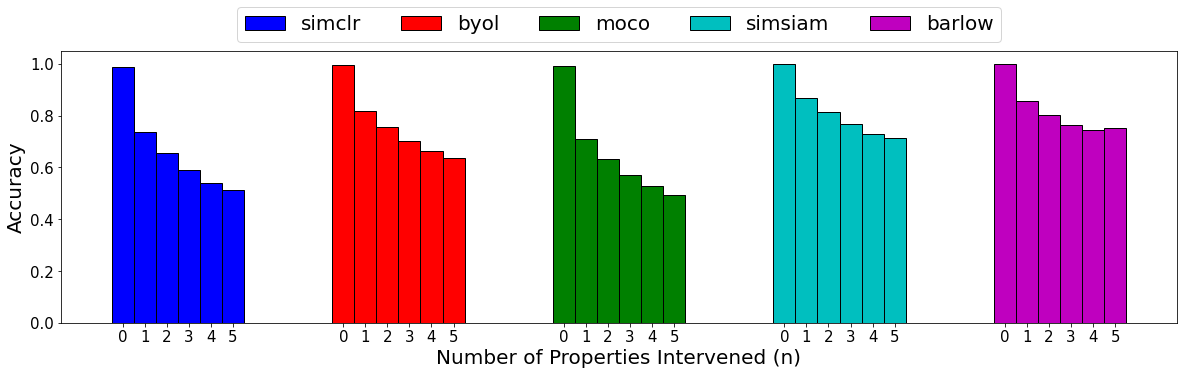}}
          \caption{Deterioration in Accuracy}
          \label{fig:child_acc}
     \end{subfigure}
     \begin{subfigure}[b]{0.5\textwidth}
          \centering
          \resizebox{\linewidth}{!}{\includegraphics{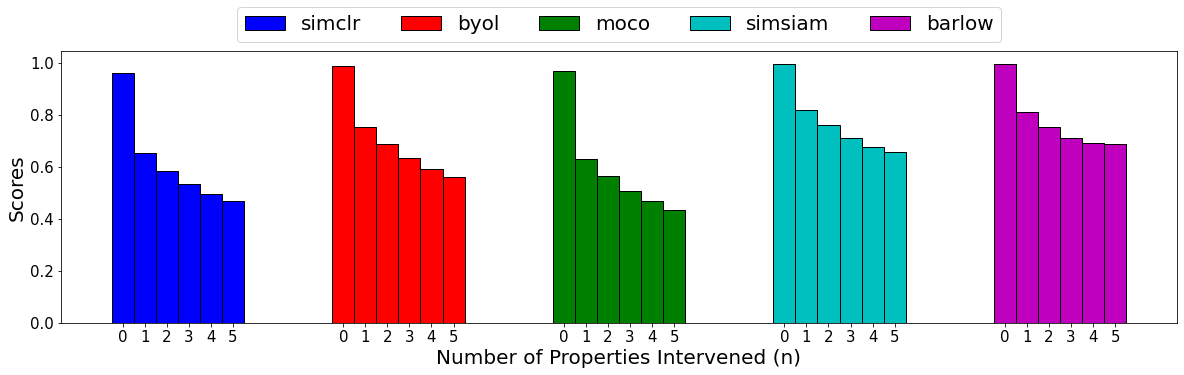}}  
          \caption{Deterioration in Score}
          \label{fig:child_score}
     \end{subfigure}
    \caption{Deterioration of unstable changing in data variables on all SSL. Only 5 children nodes are intervened.}
    \label{fig:child_causal}
\end{figure}

We also verify the same deterioration in performance when only intervening the 5 children nodes in Figure \ref{fig:causal3d}. In Figure \ref{fig:child_causal}, we observe the same deterioration when only 5 variables are sampled to exclude the extreme edge(s).

We also visualize the latent shift between stable and unstable examples via T-SNE\cite{van2008visualizing}. We employed prediction scores and accuracy to quantitatively demonstrate the effectiveness and value of our proposed solutions, ensuring the soundness of our research and effectively showcasing their benefits.

\begin{figure*}[htbp]
    \centering
    \includegraphics[scale=0.35]{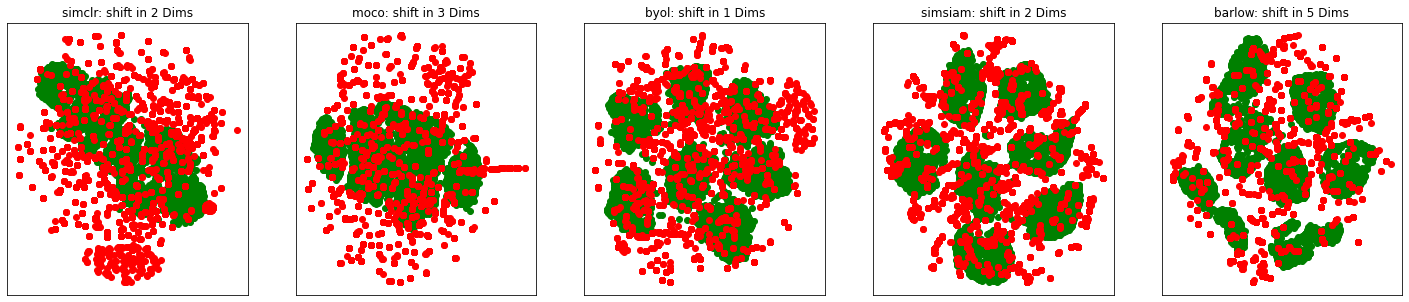}
    \caption{\textcolor{green}{Stable latents} are well clustered. \textcolor{red}{Unstable latents} are scattered randomly around stable clusters.}
    \label{fig:clusters}
\end{figure*}

\section{ImageNet} \label{appendix:imagenet}
\textbf{ObjectNet} is a large crowdsourced test set for object recognition that includes controls for object
rotations, viewpoints, and backgrounds. Objects are posed by workers in their own homes in natural
settings according to specific instructions detailing what object class they should use, how and where
they should pose the object, and where to image the scene from. Every image is annotated with these
properties, allowing us to test how well object detectors work across these conditions. Each of these
properties is randomly sampled leading to a much more varied dataset. There are 313 ObjectNet classes with 113 of them overlapping with the ImageNet classes. With each controlled variable, the changes in the variable poses challenges to identify the objects correctly due to the very unusual shift. 

\begin{figure}[htbp]
    \centering
    \includegraphics[scale=0.3]{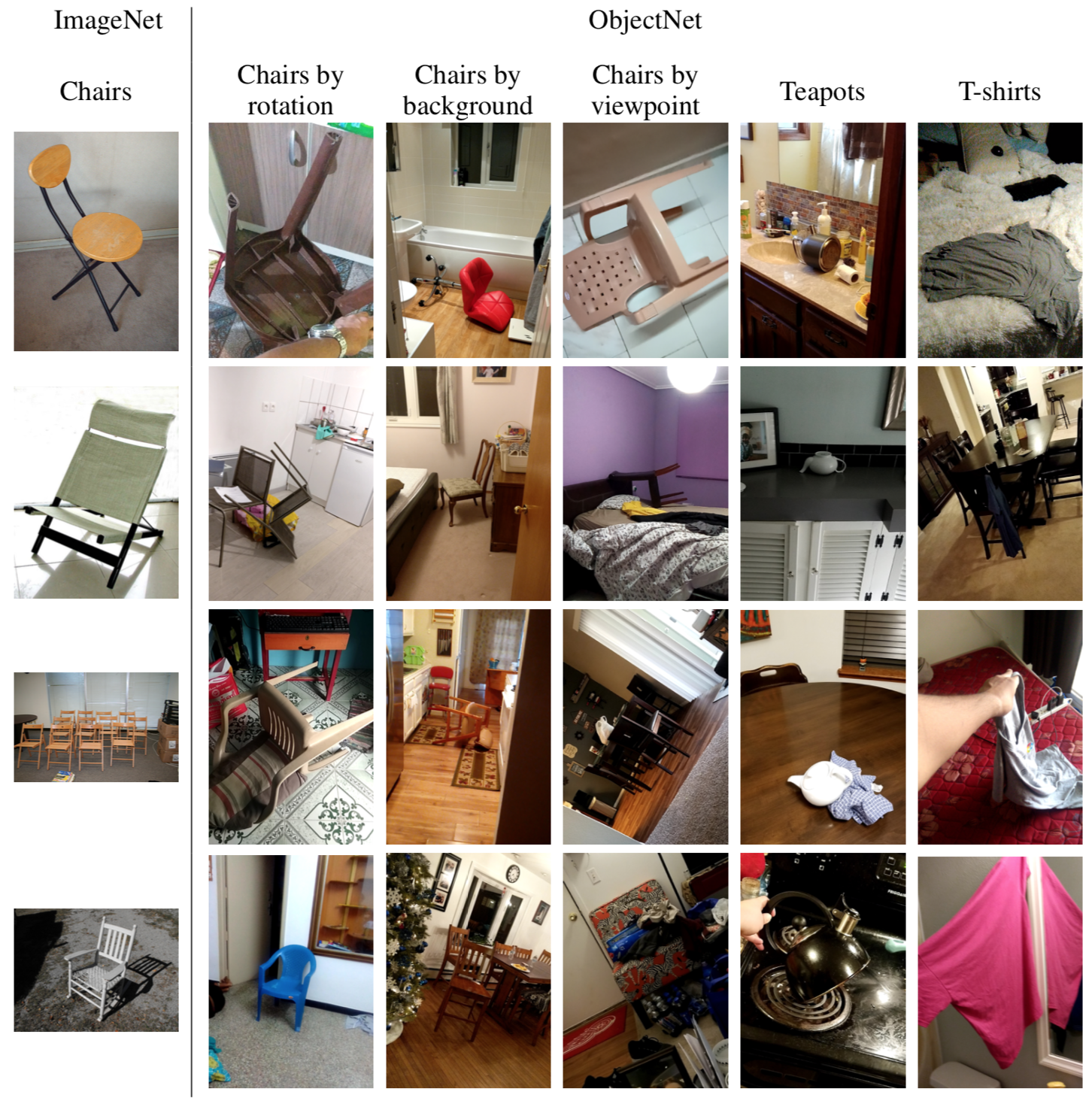}
    \caption{Illustration of ObjectNet controlling data variables.}
    \label{fig:objectnet}
\end{figure}

Starting from ImageNet \textbf{Stylized ImageNet} is constructed by stripping every single image of its original texture and replacing it with the style of a randomly selected painting through AdaIN style transfer. The original objective for Stylized-ImageNet is to help the network to learn more about shapes, and less about local textures. However, we regards this shift in the appearance as a change in the data variables(s). Since the stylized images appear drastically different to natural images, we assume this shift is very hard to counter in the representation space due to entangled transformations.

\begin{figure}[htbp]
    \centering
    \includegraphics[scale=0.15]{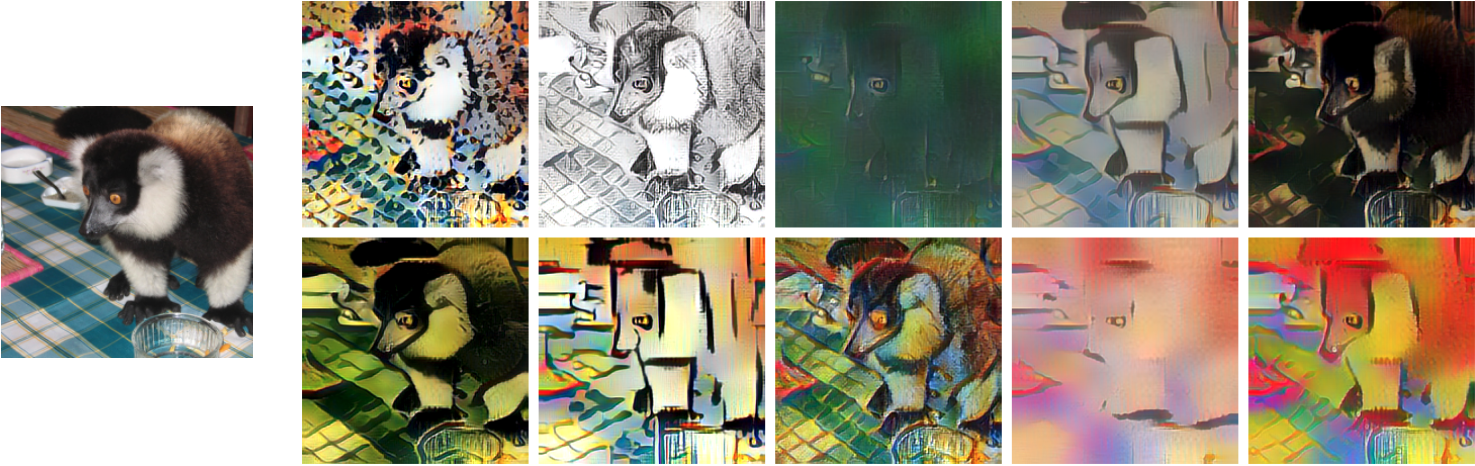}
    \caption{Examples of Stylized ImageNet}
    \label{fig:stylized2}
\end{figure}

\textbf{Synthetic Data} follows synthetic procedure in \cite{djolonga2020robustness} where object is masked on a background at a location with a rotation angle. Foreground object masks are cropped from OpenImages \cite{kuznetsova2020open}. The object classes that overlap with ObjectNet classes are selected, and each class is selected with 10 object masks at highest area and not truncated by any other objects. The backgrounds are sampled from \textit{pexel.com} with the same set of 867 backgrounds used in \cite{djolonga2020robustness}. Initially there are three data variables to control with: \textit{background}, \textit{rotation}, \textbf{location}. The object class is randomly selected at each sythesizing step. At each training step, the other two variables are randomly fixed while the target variable is randomly selected across 10 values. This results in 10 images to compare with each other. For the special data variable, \textit{texture}, we change the style of the selected object masks to different textures in \textit{Describable Textures Dataset} \cite{cimpoi14describing} based on formulation used in \textbf{Stylized ImageNet}. With all other three variables randomly fixed, the object mask with 10 different textures are sampled.

\begin{figure}[htbp]
    \centering
    \begin{subfigure}[b]{0.5\textwidth}
          \centering
          \resizebox{\linewidth}{!}{\includegraphics{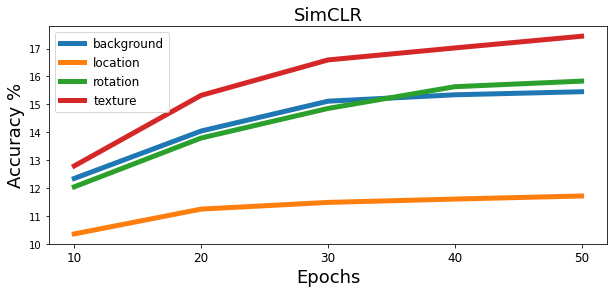}}
          \label{fig:s2_simclr}
     \end{subfigure}
     \begin{subfigure}[b]{0.5\textwidth}
          \centering
          \resizebox{\linewidth}{!}{\includegraphics{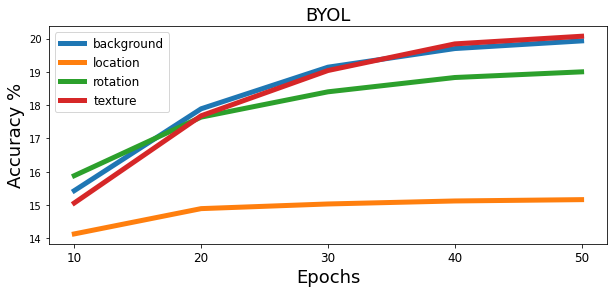}}  
          \label{fig:s2_byol}
     \end{subfigure}
     \begin{subfigure}[b]{0.5\textwidth}
          \centering
          \resizebox{\linewidth}{!}{\includegraphics{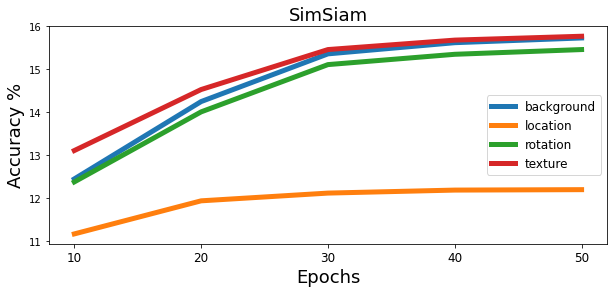}}  
          \label{fig:s2_simsiam}
     \end{subfigure}
    \caption{Training longer in \textbf{Stable Inference Mapping} can improve the performance. But the improvement saturates after around 30 epochs and the improvement becomes less significant.}
    \label{fig:s2_objectnet}

\end{figure}

We carry the experiment on \textbf{Stable Inference Mapping} to 50 epochs and observe the saturation of the improvement (Figure \ref{fig:s2_objectnet}). To further improve the performance, more integrated interventions should be applied to make $\mathbf{F}$ more robust to shifts in the data variables.

